%% file: neurips_2024.tex
\documentclass{article}

\usepackage[preprint]{neurips_2024}

\usepackage[utf8]{inputenc} 
\usepackage[T1]{fontenc}    
\usepackage{hyperref}       
\usepackage{url}            
\usepackage{booktabs}       
\usepackage{amsfonts}       
\usepackage{nicefrac}       
\usepackage{microtype}      
\usepackage{xcolor}         
\usepackage[nolist]{acronym}
\usepackage{amsmath}
\usepackage{amsthm}
\usepackage{amssymb}
\usepackage{pifont} 
\usepackage{graphicx}

\newcommand{\vect}[1]{\mathbf{#1}}
\newcommand{\rot}[1]{\left[#1\right]^\alpha}
\newcommand{\cmark}{\ding{51}}%
\newcommand{\xmark}{\ding{55}}%
\newcommand{\myparagraph}[1]{
     \textbf{#1}
     }

\theoremstyle{definition} 
\newtheorem{definition}{Definition}[section] 
\newtheorem{lemma}{Lemma}[section]
\newtheorem{theorem}{Theorem}[section]
\title{Harmformer: Harmonic Networks Meet Transformers for Continuous Roto-Translation Equivariance}

\setcitestyle{numbers,open={[},close={]},citesep={,}}

\author{%
  Tomáš Karella, Adam Harmanec, Jan Kotera, Jan Blažek, Filip Šroubek  \\
  Department of Image Processing \\
  Institute of Information Theory and Automation \\
  Czech Academy of Sciences \\
  \texttt{karella@utia.cas.cz} \\
}
\begin{acronym}[TDMA]
\acro{he}[HE]{Harmonic Equivariance}
\acro{ge}[GE-ViT]{$E(2)$-Equivariant Vision Transformer}
\acro{gsa}[GSA-Nets]{Group Equivariant Self-Attention Networks}
\acro{vit}[ViT]{Visual Transformer}
\acro{sa}[SA]{Self-Attention}
\acro{hnets}[H-Nets]{Harmonic Networks}
\acro{cnns}[CNNs]{Convolutional Neural Networks}
\acro{gcnn}[G-CNNs]{Group Equivariant Convolutional Neural Networks}
\end{acronym}

\begin{document}

\maketitle
\begin{abstract}
\acl{cnns} exhibit inherent equivariance to image translation, leading to efficient parameter and data usage, faster learning, and improved robustness. The concept of translation equivariant networks has been successfully extended to rotation transformation using group convolution for discrete rotation groups and harmonic functions for the continuous rotation group encompassing $360^\circ$.  We explore the compatibility of the \acl{sa} mechanism with full rotation equivariance, in contrast to previous studies that focused on discrete rotation. We introduce the Harmformer, a harmonic transformer with a convolutional stem that achieves equivariance for both translation and continuous rotation. Accompanied by an end-to-end equivariance proof, the Harmformer not only outperforms previous equivariant transformers, but also demonstrates inherent stability under any continuous rotation, even without seeing rotated samples during training. 
\end{abstract}

\begin{figure}[h]
  \centering
  \includegraphics[width=\textwidth]{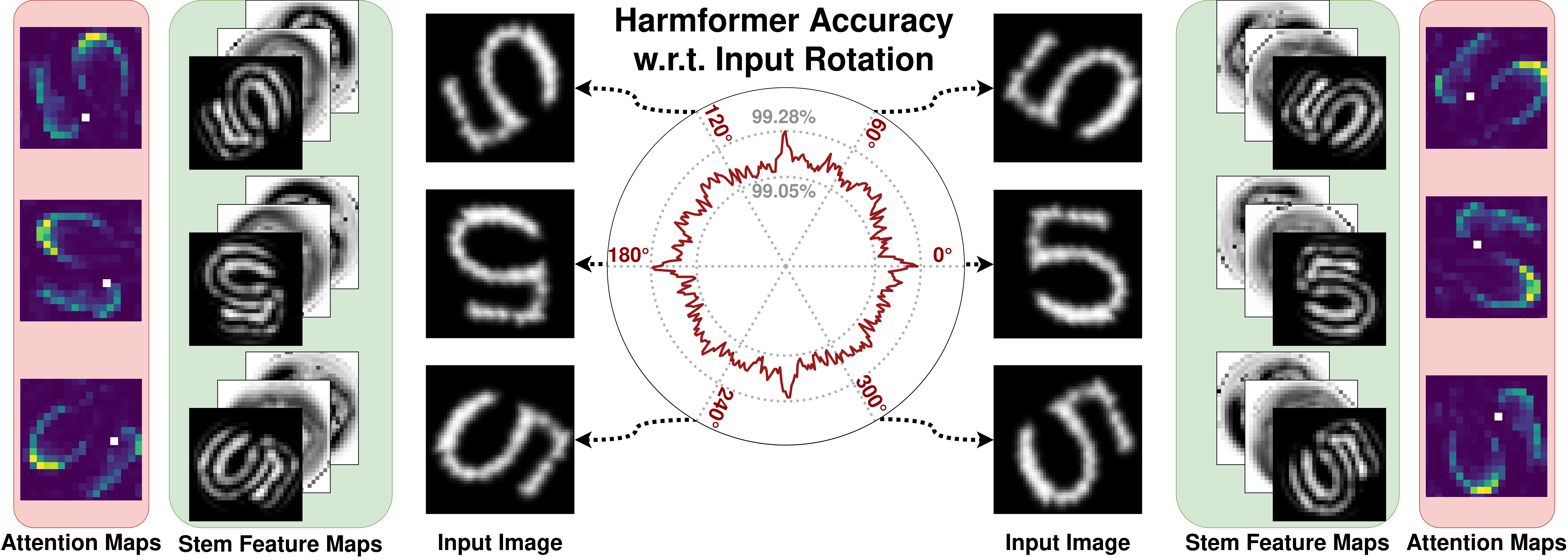}
  \caption{Equivariance of the Harmformer feature and attention maps in response to rotation of the input image: While the maps themselves are rotated, the magnitudes in the maps remain the same.}  
  \label{fig:intro}
\end{figure}
\section{Introduction}
A key strength that positions \ac{cnns} \citep{lecun1998} as a superior architecture for computer vision tasks is the weight sharing across the spatial domain. This design ensures that CNN feature maps retain their values as the input is translated, only being shifted according to the input. Formally known as translation equivariance, this property provides \ac{cnns} with inherent robustness and efficiency in managing translations. Equivariance can be extended to other groups of transformations, such as rotation, scaling, or mirroring. The advantage of equivariant models is that they ensure a tractable response of the model to the transformation of the input. As a result, the model can eliminate the effects of the transformations and produce predictions that are invariant to them. For instance, to achieve translation invariance in conventional CNNs, the feature maps are commonly aggregated by global average pooling before the classification layer.

\ac{gcnn} \citep{cohen2016} show that \ac{cnns} can be modified to become equivariant to any discrete transformation group, such as rotation by a discrete set of angles. An extension to continuous rotation and translation group was introduced by \citet{worrall2017}. The authors proposed \ac{hnets}, which restrict the convolution filters to a family of harmonic functions ideal for expressing full rotation equivariance. Both approaches improve the generalization and efficiency of training for the chosen group, similarly to how \ac{cnns} benefit from translation equivariance. For example, rotation equivariant networks are well suited for object detection in aerial imagery because such images lack natural orientation and equivariant networks inherently accommodate all rotations. Beyond aerial imagery \citep{han2021,ding2019}, equivariant \ac{cnns} are effective in many other applications, such as microscopy \citep{chidester2019a, chidester2019b}, histology \citep{graham2020}, and remote sensing \citep{cheng2019}.

With the adoption of transformer architectures in computer vision, the \acl{sa} mechanism has also been integrated into equivariant networks \citep{fuchs2020, romero2020, hutchinson2021}. Equivariant transformers are gaining importance especially in domains such as graph-based structures (e.g. molecules) \citep{agashov2024, liao2024, liao2023, tholke2022}, vector fields \citep{assaad2023}, manifolds \citep{he2021}, and generic geometric data \citep{haan2024, brehmen2023}. In the 2D domain, \citet{romero2021} proposed a transformer equivariant to discrete rotation and translation groups by using the principle of \ac{gcnn} in the positional encoding of the \ac{sa}. The formulation was further improved by \citet{xu2023}. In both cases, the computational complexity of the equivariant \ac{sa} increases quadratically with the number of angles in the considered rotation group, which limits the model angular resolution. Equivariance to continuous rotation presents a versatile solution.

In this paper, we introduce Harmformer, the first vision transformer capable of achieving continuous 2D roto-translation equivariance. The name is derived from circular harmonics~\citep{freeman1991} which provide the equivariance property preserved throughout the architecture. To ensure computational efficiency, our network starts with an equivariant convolutional stem based on Harmonic networks~\citep{worrall2017}, where we redesign the key components, such as activations, normalization layers, and introduce equivariant residual connections. The stem output is divided into equivariant patches, which are then passed to the transformer. Alongside a novel self-attention \ac{sa} mechanism, we introduce layer normalization and linear layers to guarantee end-to-end equivariance. The equivariance property allows Harmformer to remove the effect of roto-translation just before classification, preserving all relevant information at earlier stages (see Fig.~\ref{fig:intro}).

Through experimental validation, we show that Harmformer surpasses all previous discrete equivariant transformers \citep{romero2021, xu2023} on established benchmarks \citep{larochelle2007, bejnordi2017, veeling2018}. It also outperforms earlier invariant models \citep{karella2023, hwang2021, khasanova2017} on classification tasks where the model is trained solely on non-rotated data. 

\section{Related Work}
We review the three foundational concepts from prior research that Harmformer builds upon: the \ac{sa} mechanism, equivariant convolution networks, and transformers with a convolutional stem stage. Additionally, we discuss other equivariant transformer architectures.

\myparagraph{Visual Self-Attention}
The well-known \ac{sa} mechanism originates from natural language processing \citep{vaswani2017} and is widely used in computer vision since the publication of the \ac{vit} \citep{dosovitskiy2021}.  Transformers, unlike \ac{cnns}, exhibit larger model capacities but require substantial amounts of data and have quadratic complexity with respect to input size. Transformers closely related to Harmformer include CoAtNet \citep{dai2021} and, more specifically, $\text{ViT}_p$ \citep{xiao2021}. These architectures begin with a convolution stem to downscale the input and thereby reduce the computational complexity of the subsequent application of \ac{sa}. However, these architectures are not equivariant to roto-translation.

\myparagraph{Equivariant Convolutions}
Since the publishing of the \ac{gcnn}~\citep{cohen2016}, the concept of equivariant convolutional networks has expanded across various modalities and transformation groups. In 2D, these transformations include rotation \citep{worrall2017, weiler2018b}, scaling \citep{worrall2019, sosnovik2020, rahman2023}, and general $E(2)$ transformations \citep{weiler2019}. In 3D, applications cover $SO(3)$ transformations in volumetric data \citep{weiler2018a, worrall2018} and point clouds \citep{bekkers2024}, as well as spherical \ac{cnns} \citep{cohen2018}. Equivariant networks are also applied to graphs \citep{satorras2021} and non-Euclidean manifolds \citep{weiler2021}. Harmformer builds on and extends the \ac{hnets} published by \citet{worrall2017}, which are purely convolutional networks equivariant to continuous $360^\circ$ rotation. In our implementation of \ac{hnets}, we incorporate the improvements introduced in H-NeXt \citep{karella2023}.

\myparagraph{Equivariant Transformers}
As previously mentioned, equivariant networks have integrated the \ac{sa} mechanism in various domains, including 3D graphs and point clouds using irreducible representations \citep{fuchs2020, liao2023, liao2024}, operations on Lie algebras \citep{hutchinson2021}, and general geometric data using geometric algebras \citep{haan2024, brehmen2023}. Particularly relevant to our work are the planar roto-translation equivariant transformers, such as \ac{gsa} \citep{romero2021} and \ac{ge}, which reformulate relative positional encoding to construct equivariant transformers. However, \ac{gsa} and \ac{ge} operate only on discrete rotation groups such as $\left\{ k \frac{\pi}{2} \mid k \in \mathbb{Z} \right\}$, where finer angular sampling substantially increases the computational complexity.  

\section{On Equivariance in Vision Transformers}
\label{sec:equivariance}
We analyze the roto-translation equivariance of the \ac{vit} architecture, a well-known representative of vision transformers. First, we formalize the notion of equivariance. Intuitively, a function $f$ is equivariant to a transformation $a_g$ if the transformation and the function commute, $f(a_g(x)) = a_g(f(x))$. For example, processing a rotated input image has the same effect as directly rotating the features of the unrotated image. In practice, such a definition would be too restrictive. The function $f$ (layer or network) typically has a different domain and codomain, so the transformation may act differently on each. The core idea remains the same: the model response to the input transformation is predictable. To formally define equivariance, we draw upon the seminal work of \citet{cohen2016} or the more recent one formulated by \citet{weiler2023}.
\begin{definition}[Equivariance]
\label{def:equivariance}
A function $f: X \rightarrow Y$ (a whole network or a single layer) is called group equivariant with respect to a group $G$ if for every element $g$ in $G$, represented by a linear map $a_g: X \to X$, there exists a corresponding linear map $b_g: Y \to Y$ such that the following holds:
\begin{equation}\label{eq_equivariance_def}
f(a_g(x)) = b_g(f(x)) \quad \text{for all } x \in X \text{ and } g \in G.
\end{equation}
\end{definition}
A composition $f_2(f_1(x))$ of two equivariant functions $f_1: X \rightarrow Y$ and $f_2: Y \rightarrow Z$ is equivariant. We call invariance a special case of equivariance when $b_g$ is the identity for all $g$ in $G$.

\myparagraph{Self-Attention}
A key mechanism that distinguishes transformers from previous architectures is the \ac{sa} layer \citep{vaswani2017}. Before discussing the properties of \ac{sa}, let us formally define it. 
\begin{definition}[Self-Attention]
\label{def:self-attention}
Given an input matrix $Y \in \mathbb{R}^{n \times d}$, where each row of $Y$ represents a feature vector of dimension $d$, usually called a patch. The matrices $Q$ (queries), $K$ (keys), and $V$ (values) are computed as linear projections of $Y$:
\begin{align}
\label{eq:qkv}
    [Q, K, V] &= [Y W_q, Y W_k, Y W_v ]  & W_{q,k,v} \in \mathbb{R}^{d \times d_h},
\end{align}
where $d_h$ is the dimension within the \ac{sa} layer. The output of the self-attention layer, $\text{SA}(Y)$, is a weighted sum of the vectors in $V$, where the weights are defined as the softmax-normalized pairwise similarity scores between the vectors in $Q$ and $K$:
\begin{align}
\label{eq:att}
    A &= \text{softmax}(QK^T / \sqrt{d_h}) & A \in \mathbb{R}^{n \times n}, \\
\label{eq:av}
    \operatorname{SA}(Y) &= AV.
\end{align}
In practice, \ac{sa} is typically extended to Multi-Head \acl{sa} (MSA), in which multiple \ac{sa} layers with different embedding matrices $W_{(k, v, q)}$ are computed in parallel and then combined.
\end{definition}

The construction of \ac{sa} implies its well-known property, in the literature often referred to as permutation invariance \citep{yang2023}. According to Def.~\ref{def:self-attention}, it is more accurate to call the \ac{sa} layer permutation equivariant rather than permutation invariant. For if we change the order of the rows in $Y$, the $\operatorname{SA}(Y)$ remains the same except for the same change in the order of its output rows. 

\myparagraph{What makes \ac{vit} non-equivariant?} As rotation and translation are special cases of permutation, the permutation equivariance of \ac{sa} might suggest the roto-translation equivariance of the whole \ac{vit}. However, the permutation-equivariance of \ac{sa} holds at the patch level and not at the pixel level, where translation or rotation takes place. In the initial stage of \ac{vit}, before the first \ac{sa} layer, the image is split into $n$ non-overlapping patches of fixed size, typically 16$\times$16 pixels. These are linearly transformed and flattened to form the rows of the  input matrix $Y$ of the first \ac{sa} layer. This patch-wise operation breaks the direct rotation (or translation) equivariance of \ac{vit} at the image level, because for an image rotation $a_g$ in Eq.~\eqref{eq_equivariance_def} corresponding to an angle $g$ from the rotation group $G$, there is no $b$ acting on the patches that can be expressed as a rotation $b_g$ by $g$; the same holds for translation.

A solution typically used by previous equivariant approaches \citep{romero2021, xu2023} is to consider pixel-level ``patches'' of size 1$\times$1 pixel. Then the image rotation is equivalent to patch-level permutation, and the corresponding transformer model remains equivariant, assuming that interpolation errors and boundary effects are minimal. This approach, however, has two major drawbacks. As seen from Eq.~\eqref{eq:att}, the \ac{sa} has a quadratic complexity with respect to the number of patches $n$, so operating on a pixel grid (as opposed to 16$\times$16) incurs an almost $10^5$ penalty factor in memory requirements and correspondingly increases the processing time. To mitigate this, \ac{gsa} and \ac{ge} reduce complexity by using local \ac{sa} \citep{prajit2019} that restricts the attention field to the 7$\times$7 neighborhood of the patch. The second drawback is that the local self-attention in the first layers is not very informative, because nearby pixels are usually highly correlated.

\myparagraph{Position Encoding}
During construction of the input matrix $Y$ for the first \ac{sa} layer, the patches are also given absolute position encoding that provides information about their locations. This breaks equivariance as the patches of a transformed image will receive different encoding compared to their counterparts in the original image. Equivariant transformers \citep{romero2021, xu2023} replace the absolute encoding with circular relative encoding, similar to iRPE introduced by \citet{wu2021}. 

In Harmformer, we address these challenges with a convolutional stem stage that initially reduces spatial dimensions and extracts high-level features. Subsequently, we create 1$\times$1 patches from these high-level features and process them by the \ac{sa} layers. To maintain spatial correspondence among the patches while ensuring equivariance, Harmformer also uses circular relative position encoding.

\section{Harmonic Convolutions and Equivariance to Continuous Roto-Translation}
To understand the equivariance property of Harmformer, it is essential to understand the concept of harmonic convolutions introduced in \ac{hnets} \citep{worrall2017}, as they are employed in the stem and affect the subsequent transformer layers. The main difference from the traditional \ac{cnns} is that the convolution filters based on circular harmonic functions are specifically designed to encode rotational symmetries. The filters are defined as follows.

\begin{definition}[Harmonic Filter]
\label{def:harmonic-filter}
A harmonic filter $W_m: \mathbb{R}^2 \rightarrow \mathbb{C}$ parameterized by a rotation order $m$ is given by:
\begin{equation}
W_m(r, \theta) = R(r) e^{i (m \theta + \beta)},
\end{equation}
where $(r, \theta)$ are polar coordinates. Here, $R: \mathbb{R} \rightarrow \mathbb{R}$ is a learnable radial function and $\beta \in \mathbb{R}$ is a learnable phase shift. The rotation order $m$ is a parameter that determines the filter symmetry.
\end{definition}

As translation equivariance is inherently provided by convolution, we will focus solely on rotation in the following discussion and denote the rotation operator by $\rot{\cdot}$, where $\alpha$ is the angle of rotation. Let us look in detail at how the rotation applied to the input affects feature maps generated by harmonic convolution. The \ac{hnets} features are represented as complex values in polar form.
\begin{lemma}[Harmonic Convolution Property]
\label{lemma:hconv}
Let $I$ be an input image and $W_{m_1}$ a harmonic filter. Under image rotation by angle $\alpha$, convolution of $I$ with $W_{m_1}$ is given by:  
\begin{equation}
\rot{I} \circledast W_{m_1} = e^{i m_1 \alpha} \rot{I \circledast W_{m_1}}\,.
\end{equation}
This equation shows that rotating the image only results in a phase shift of the feature values, while the spatial coordinates are rotated accordingly. This property also holds for subsequent convolution layers. If the first feature map is denoted as $F_{m_1}(\rot{I})$, then convolution with another harmonic filter $W_{m_2}$ is given by:
\begin{equation}
F_{m_1}(\rot{I}) \circledast W_{m_2} = e^{i (m_1 + m_2) \alpha} \rot{F_{m_1}(I) \circledast W_{m_2}}\,.\label{eq:mixing}
\end{equation}
\end{lemma}
The authors of \ac{hnets} also construct activation, batch normalization, and pooling layers that preserve this property. As a result, their classifier can be independent of input rotation and translation. To remove the influence of rotation, they extract only the magnitude from the last feature map and discard the phase. To aggregate spatial information, they use global average pooling. Note that for tasks where the orientation of the object is relevant, phase can be used as no information is lost due to equivariance. 

To unify the equivariance property within the Harmformer architecture, we define \ac{he}, which is motivated by Lemma~\ref{lemma:hconv} and satisfies the general definition of equivariance (Def.~\ref{def:equivariance}). \ac{he} describes how features transform with respect to the rotation of an input image. By showing that each Harmformer layer satisfies \ac{he}, we establish the relationship between the features and the rotation of an input throughout the model.

\begin{definition}[Harmonic Equivariance -- HE]
\label{def:harmonic-equivariance}
A layer $F_m (\cdot)$ associated with a rotation order $m$ is said to be \ac{he}, if for any rotation by angle $\alpha$ and admissible input $I$, it is transformed as follows:
\begin{equation}
F_m (\rot{I}) = e^{i m \alpha} \rot{F_m (I)}.
\end{equation}
Here $\rot{F_m(I)}$ are features obtained from an unrotated input $I$ and then rotated. The phase is shifted by a multiple of the rotation angle, where the factor is given by the rotation order of the layer. The process is illustrated in Fig.~\ref{fig:backbone}a.
\end{definition}

\section{Harmformer Architecture}
\label{sec:harmformer}
\begin{figure}[h]
  \centering
  \includegraphics[width=\linewidth]{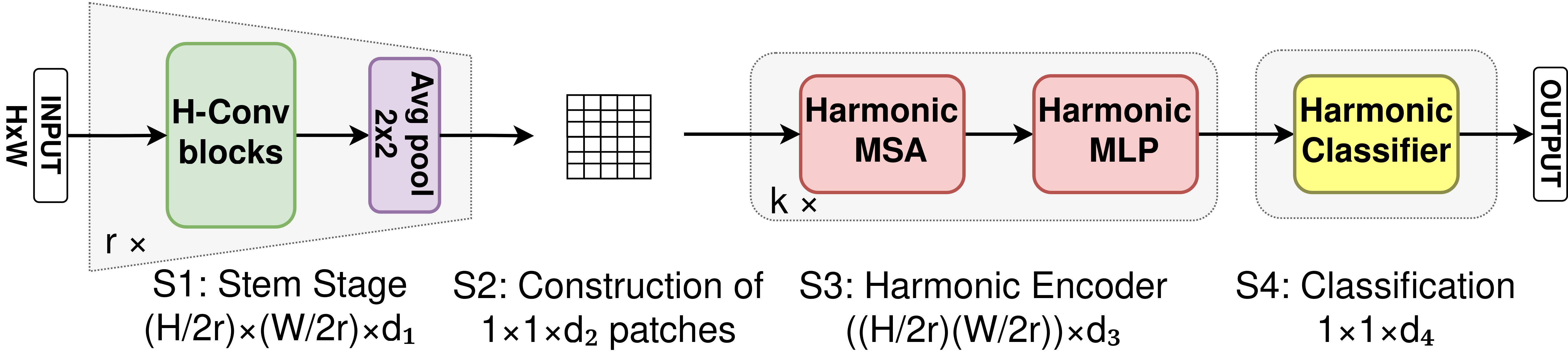}
  \caption{Overview of the Harmformer architecture, divided into four stages: S1 - downscaling the input, S2 - constructing patches from feature maps, S3 - Harmonic Encoder, and S4 - Classifier.} 
  \label{fig:architecture}
\end{figure}

The architecture of Harmformer is shown in Figure \ref{fig:architecture} and its layers will be discussed one by one. 
\ac{he} (Def. \ref{def:harmonic-equivariance}) of each layer is proved in Appendix \ref{appendix:equivariance-proofs}, demonstrating the end-to-end continuous rotation and translation equivariance. The architecture begins with a stem stage based on \ac{hnets}, which we have further improved by refining activation and normalization layers and incorporating residual connections. The stem is followed by an equivariant encoder tailored to maintain \ac{he}, and  the last component is a classifier, which takes the \ac{he} output of the encoder and computes an invariant representation for classification.

\subsection{Harmformer: S1 Stem Stage}
\label{subsec:backbone}
The main role is to prepare features for the Harmonic Encoder (S3) so that they are \ac{he} and have lower spatial resolution to keep the computational complexity of \ac{sa} manageable, as discussed in Sec. \ref{sec:equivariance}. To this end, we design the stage to comprise $r$ iterations of H-Conv blocks, followed by average pooling, as shown in Fig.~\ref{fig:architecture}. Each iteration increases the number of channels while decreasing the spatial dimension. 

\begin{figure}[h]
  \centering
  \includegraphics[width=\linewidth]{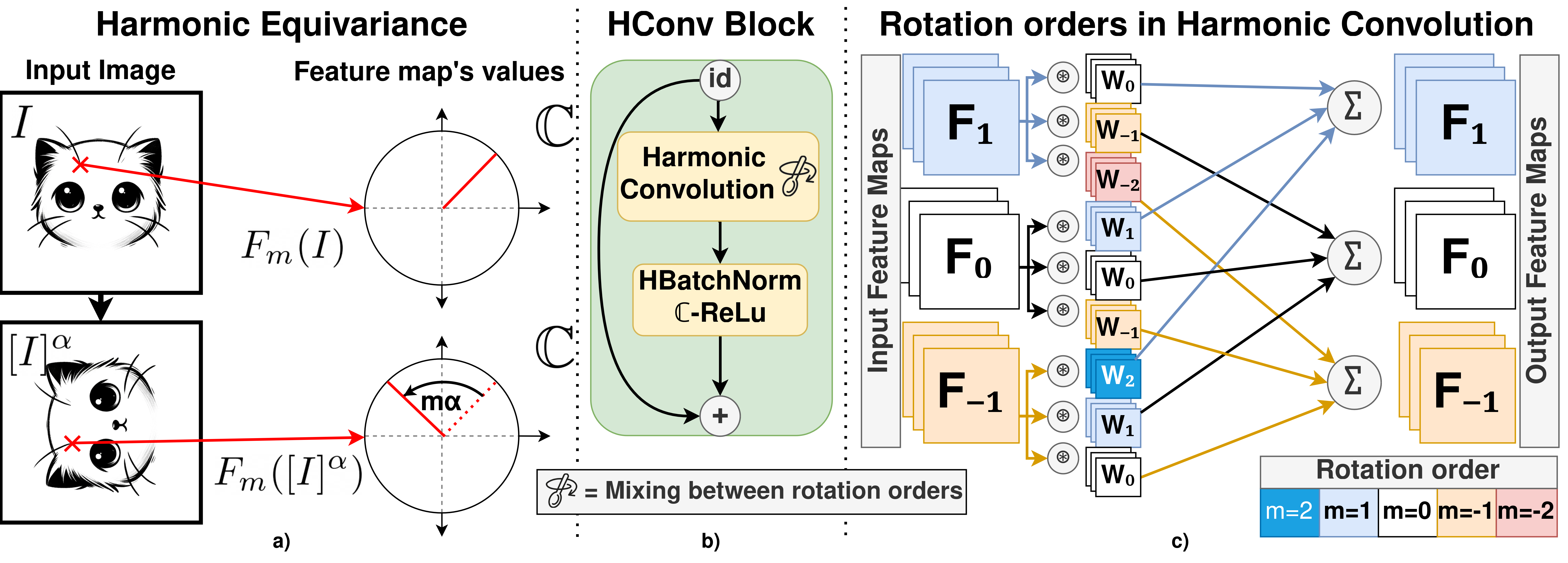}
  \caption{(a) Phase shift of \ac{he} feature values when the input is rotated; (b) Harmonic Convolution (H-Conv) Block of the stem stage; (c) Interaction of harmonic filters $W_m$ with feature maps $F_m$ within the Harmonic Convolution layer of the H-Conv Block, where $m$ is the rotation order. 
  }
  \label{fig:backbone}
\end{figure}

The stage starts with an input that formally satisfies \ac{he} for the rotation order $m=0$, expressed as $\rot{I} = e^{i 0 \alpha} \rot{I}$, followed by the first H-Conv block shown in Figure \ref{fig:backbone}b.

\myparagraph{Rotation Order Streams} The \ac{he} and the definition of Harmonic Convolution have already been detailed in Lemma \ref{lemma:hconv} and Def. \ref{def:harmonic-filter}.  An important aspect that remains to be addressed is the selection of rotation orders for the harmonic filters. In our initial convolution with the input image (often called lifting convolution), we employ harmonic filters of rotation orders $-1$, $0$, and $1$. This setup produces three streams of feature maps, each corresponding to one of these rotation orders.

Our experiments, along with the results reported in \cite{karella2023, worrall2017}, indicate that generating feature maps of higher rotation orders does not significantly improve performance but increases the computational complexity. Based on this evidence, we limit rotation orders to $-1$, $0$, and $1$.

Most layers process these streams independently and those that interact across streams are indicated in the diagrams by a "spoon" symbol, as in the case of Harmonic Convolution in Figure \ref{fig:backbone}b. 
Streams in the Harmonic Convolution block are mixed similarly as in \ac{hnets}. The proposed mixing strategy is shown in Figure \ref{fig:backbone}c and follows from the Harmonic Convolution property in \eqref{eq:mixing}, which states that
\begin{equation}
F^{out}_m = \sum_{m = m_1 + m_2} F^{in}_{m_1} \circledast W_{m_2},
\end{equation}
where $m$, $m_1$, and $m_2$ are the rotation orders of the output, input, and harmonic filter, respectively. 

\myparagraph{Layers Operating on Magnitude} 
Because rotation affects only the phase of the features leaving the magnitude untouched, element-wise functions, such as normalization or activation, operating only on magnitudes preserve the \ac{he} property. In contrast with previous \ac{hnets} \citep{karella2023, worrall2017}, we restrict the codomain of every element-wise function $f$ transforming magnitudes to non-negative numbers, $f: \mathbb{R} \rightarrow \mathbb{R}_{0}^{+}$, since negative magnitudes inadvertently flip the phase, thus violating the \ac{he} property. This consideration leads us to propose a novel normalization fused together with activation (HBatchNorm and $\mathbb{C}$-ReLu), detailed in Appendix \ref{appendix:hnormact}. Restricting the codomain and fusing the normalization with the activation has a positive impact on performance, as shown in Ablation \ref{appendix:norm}.  

\myparagraph{Residual Connection} The final stem element is the residual connection, previously unused in \ac{hnets}. Residual connections are also used within our encoder blocks. As in standard \ac{cnns}, they improve gradient flow and reduce training time. With respect to rotation orders, they process streams independently, thus preserving \ac{he} according to the following lemma:
\begin{lemma}[\ac{he} of Residual Connections]
\label{lemma:residual}
A residual connection between feature maps of the same rotation order, $F_m(I)$ and $F'_m(I)$, preserves \ac{he} property:
\begin{equation}
    F'_m(\rot{I}) + F_m(\rot{I}) = e^{i m \alpha} \rot{(F'_m(I) + F_m(I))}.
\end{equation}
\end{lemma}

\subsection{Harmformer: S2 Construction of the Patches}
\label{subsec:patches}
\begin{figure}[h]
  \centering
\includegraphics[width=\textwidth]{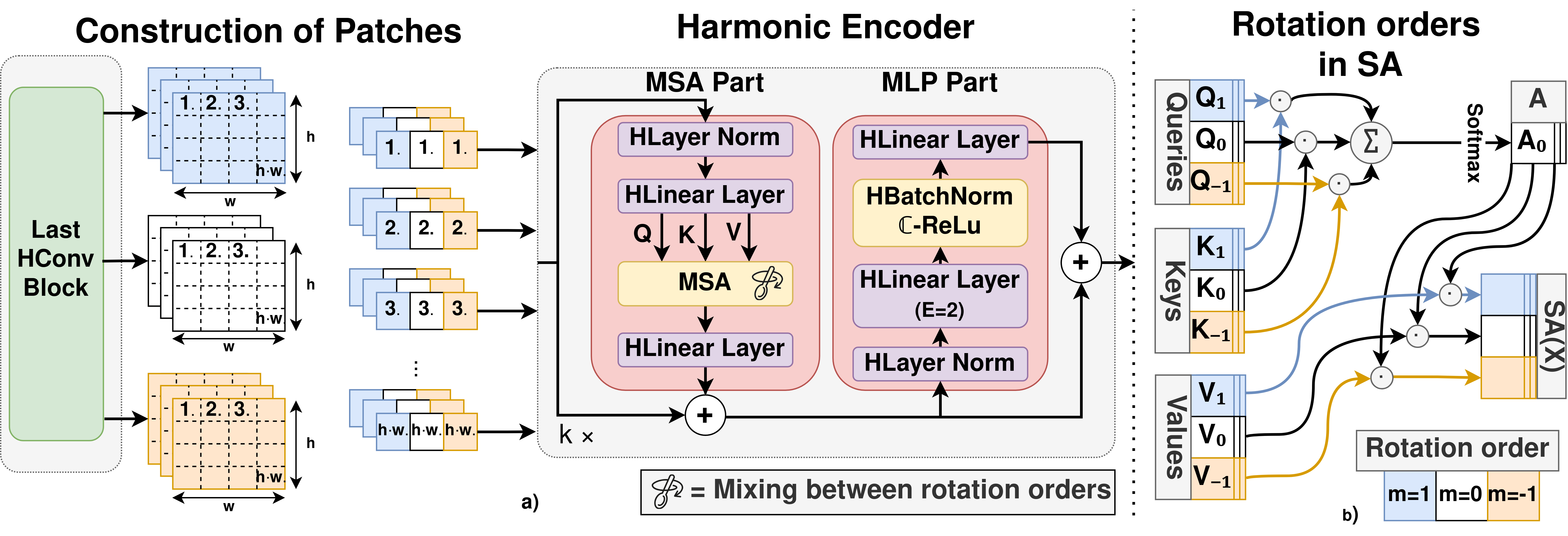}
  \caption{a) Construction of the patches (colors represent rotation orders) and the Harmonic Encoder structure. b) Diagram depicting the interaction of SA mechanisms across different rotation orders.} 
  \label{fig:encoder}
\end{figure}

To integrate the stem output with the encoder, the final stem feature maps are divided into 1$\times$1-sized patches, as illustrated in Figure \ref{fig:encoder}a. The patches are constructed separately for all three streams of rotation orders. The resulting stack of patches then comprises three matrices $F_{-1},F_{0}, F_{1} \in \mathbb{C}^{(h \cdot w) \times d}$, each representing a single rotation order, where $h$, $w$, and $d$ denote the height, width, and number of channels of the last feature maps, respectively. We keep this notation for encoder feature maps (patches), as they correspond to the stem feature maps, just reshuffled. 

Neglecting small interpolation errors, the spatial transformation of the input translates only into a permutation of the stack of patches $F_m$ as discussed in Sec. \ref{sec:equivariance}. For clarity, we use a discrete representation but it should be noted that the encoder can be modeled using a functional framework, as shown by \citet{romero2021}.  

Before feeding the \ac{sa} with patches, transformer networks typically apply a linear projection to adjust the dimension $d$.
We use a linear layer that processes the patches independently with respect to their order of rotation to preserve \ac{he}:
\begin{lemma}[\ac{he} of  Linear Layer]\label{lemma:linear}
A linear layer applied to a \ac{he} feature map $F_m(\rot{I}) \in \mathbb{C}^{(hw) \times d_{in}}$ preserves the rotation order $m$. Formally, we have: 
\begin{equation}
    F_m(\rot{I}) W = e^{m i \alpha} \rot{F_m(I) W}, 
\end{equation}
where $W \in \mathbb{C}^{d_{in} \times d_{out}}$ represents a shared weight matrix applied independently over all spatial positions of the input feature map.
\end{lemma}

\subsection{Harmformer: S3 Harmonic Encoder}
This section outlines our encoder, which is designed to preserve the \ac{he} property. The encoder is organized into several $k$ blocks, each containing Multi-Head \acl{sa} (MSA) and Multi-Layer Perceptron (MLP) components, as shown in Figure\ref{fig:encoder}a. Along with the layers presented in the previous sections, we propose a \ac{sa} mechanism and a layer normalization, both of which preserve the \ac{he}.

As the following lemma shows, the layer normalization can be adapted to satisfy \ac{he} by operating independently on the streams of rotation orders. 
\begin{lemma}[HE of Layer Norm]
\label{lemma:layer-norm}
A feature map $F_m (\rot{I}) \in \mathbb{C}^{(hw)\times d}$ with a rotation order $m$ preserves \ac{he} when normalized by its mean and standard deviation: 
\begin{equation}
   \frac{F_m(\rot{I}) - \mu}{\sigma + \epsilon} = e^{i m \alpha} \frac{\rot{F_m(I)} - \mu}{\sigma + \epsilon},
\end{equation}
where $\mu$, $\sigma$ are the sample means and standard deviations of the original feature maps computed over their spatial dimensions, respectively, and $\epsilon$ is a small constant added for numerical stability.
\end{lemma} 

\myparagraph{Self-Attention} The essential components of the encoder are MSA layers. The proposed MSA mixes features with different rotation orders. In the first step, queries, keys and values are generated for each rotation order $-1$, $0$, and $1$ independently, which preserves \ac{he} as follows from Lemma \ref{lemma:linear}. We split the \ac{sa} calculation (Eq.~\eqref{eq:att},\eqref{eq:av}) into two operations: dot product and matrix multiplication, and demonstrate their properties by the following lemmas. 

\begin{lemma}[Dot product subtracts rotation orders]
\label{lemma:dot}
    Consider two \ac{he} feature maps $Q_{m_1}(\rot{I}) \in \mathbb{C}^{(h w) \times d} $ and $K_{m_2}(\rot{I}) \in \mathbb{C}^{(h w) \times d} $ that represent queries and keys, respectively. The dot product of these feature maps is \ac{he} and has the rotation order $m_1 - m_2$. Formally, we have:
    \begin{equation}
        Q_{m_1}(\rot{I}) \overline{K_{m_2}(\rot{I})^T} = e^{i (m_1 - m_2) \alpha} \rot{Q_{m_1}(I)\overline{K_{m_2}(I)^T}},
    \end{equation}
    where $\overline{K_{m_2}(\rot{I})^T}$ denotes the complex conjugate transpose of $K_{m_2}(\rot{I})$.
\end{lemma}
\begin{lemma}[Matrix multiplication sums rotation orders]
\label{lemma:matmul}
    Consider a \ac{he} feature map $A_{m_1}(\rot{I}) \in \mathbb{C}^{(hw) \times (hw)}$ representing an attention matrix and HE feature map $V_{m_2}(\rot{I}) \in \mathbb{C}^{(hw) \times d}$ representing values. The result of their matrix multiplication is HE with a rotation order $m = m_1 + m_2$:
    \begin{equation}
        A_{m_1}(\rot{I}) V_{m_2}(\rot{I}) = e^{i (m_1 + m_2) \alpha} \rot{A_{m_1}(I) V_{m_2}(I)}.
    \end{equation}
    where $\rot{A_{m_1}(I)}$ and $\rot{V_{m_2}(I)}$ are feature maps created from unrotated $I$ and rotated afterwards.
\end{lemma}

After these operations, the relative circular encodings \citep{wu2021} are added to the result of the dot product before it undergoes the softmax activation. The Harmformer softmax operates only on magnitudes and its codomain is within $\mathbb{R}_0^{+}$ to avoid breaking \ac{he}.

We have shown that the dot product between queries $Q_{m_1}$ and keys $K_{m_2}$ results in a rotation order of $m = m_1 - m_2$. Similarly, matrix multiplication between the attention matrix $A_{m_1}$ and values $V_{m_2}$ yields a feature map with a rotation order of $m = m_1 + m_2$. The final task is to combine these rotation orders to produce output feature maps with the same number of rotation orders as the input feature maps, i.e. -1, 0, and 1.

\myparagraph{Mixing Orders in MSA} Since there are multiple strategies for combining rotation orders, we explored several of them and provide details on other configurations in Ablation \ref{appedix:mixing}. The optimal approach, according to our experiments, is shown in Figure \ref{fig:encoder} and involves:

\begin{enumerate}
    \item \textbf{Dot Product Calculation:} The dot product is computed only between the same rotation orders, separately. According to Lemma \ref{lemma:dot}, this results in three feature maps with the rotation order $0$ and dimension $\mathbb{C}^{(hw) \times (hw)}$.
    \item \textbf{Attention Matrix Formation:} 
    These results are summed to form a single matrix of rotation order $0$. A softmax function is then applied to form a single attention matrix $A_0 \in \mathbb{C}^{(hw) \times (hw)}$, preserving the rotation order $0$, because softmax function operates only on magnitudes and outputs non-negative numbers. 
    \item \textbf{Self-Attention Output:}  Finally, the self-attention output is produced by matrix multiplication of the attention matrix (rotation order zero) with the values of each rotation order. This process results in a triplet of outputs with the target rotation orders $\left(-1, 0, 1\right)$.
\end{enumerate}

Other layers, such as linear layers and residual connections, have been introduced in previous sections. By stacking \ac{he} layers, the last feature map coming from our encoder will maintain the \ac{he} property. 

\subsection{Harmformer: S4 Classification}
Spatial position and orientation are generally redundant for classification tasks, except when classifying directional objects such as arrows. In the final stage, we remove this redundant information from the feature maps and produce an invariant feature vector. The feature maps entering the classification stage form a matrix of the shape $\mathbb{C}^{3 \times n \times d}$.
To aggregate over different rotation orders, we keep only the magnitude, resulting in $\mathbb{R}^{n \times 3d}$. The spatial information is then eliminated by applying global average pooling over the dimension $n$ (patches), reducing the shape to $\mathbb{R}^{3d}$. The final feature vector, which is roto-translation invariant, is processed by a single linear layer for classification.

\section{Experiments}
\label{sec:experiments}
To validate the properties of Harmformer, we conducted experiments on four benchmarks listed in Table \ref{tab:benchmarks}. For detailed experimental configurations and an ablation study of architectural modifications, see Appendices \ref{appendix:experiments} and \ref{appendix:ablation}, respectively. Addition segmentation experiment is included in Appendix \ref{sec:segmentation}.

\begin{table}[h]
  \caption{Dataset overview, detailing sizes and whether training and test sets contain rotated images.}
  \label{tab:benchmarks}
  \centering
  \begin{tabular}{lcccccc}
    \toprule
    Dataset Name   & Sample Size & Train/Test/Val. Size & Rot. Train/Test & Ref. & Scenario \\
    \midrule
    mnist-rot-test   & $28\times28\times1$ & 50k / 10k / 10k & \xmark/\cmark & \citep{karella2023} & 1 \\
    cifar-rot-test & $32\times32\times3$ & 42k / 10k / 8k  & \xmark/\cmark & \citep{karella2023} & 1\\
    rotated MNIST & $28\times28\times1 $& 10k / 2k / 50k & \cmark/\cmark & \citep{larochelle2007} & 2 \\ 
    PCam & $96\times96\times3$ & 262k / 32k / 32k & \xmark/\xmark & \citep{bejnordi2017, veeling2018} & 2 \\ 
    \bottomrule
  \end{tabular}
\end{table}

\myparagraph{Model Architecture} 
The models are designed to match the number of parameters of the previous state-of-the-art models while maintaining the same overall architecture. Depending on the benchmark, the stem stage consists of $2$-$4$ blocks to reduce resolution, followed by $3$-$4$ harmonic encoder blocks. To ensure that the equivariant properties emerge  from the architecture, we avoid any data augmentation. Consistent with H-NeXt \citep{karella2023}, the inputs are initially upscaled by a factor of two to mitigate interpolation errors.

\myparagraph{Invariance Benchmarks} In the first scenario, we verify the equivariance of Harmformer by training the model exclusively on upright (non-rotated) data and testing it on randomly rotated data; the first two datasets in Tab.~\ref{tab:benchmarks}. Since the model is trained only on upright images, any equivariant properties must arise purely from the model design, not from the training data. 

We outperform previous methods on both datasets, as shown in Tables \ref{tab:mnist-rot-test} and \ref{tab:cifar10-rot-test}. Harmformer improves the robustness to rotation, as we discuss further in Section~\ref{appendix:angle-stability}, which partially explains the performance gain on mnist-rot-test. Stability under rotation is less enhanced on cifar-rot-test (Sec.~\ref{appendix:angle-stability}), and the superior results there are likely due to the higher model capacity of the transformer architecture, which improves the overall detection performance.

\myparagraph{Equivariance Benchmarks} In the second scenario, we compare the performance of the Harmformer on established equivariance benchmarks for roto-translation where there is no significant distribution shift between the training and test sets, either containing rotated samples or not. This evaluation assesses how our method stacks up against previous equivariant transformers that are equivariant to discrete rotation and translation. Tables \ref{tab:rot-mnist} and \ref{tab:pcam} show the top results on the rotated MNIST and PCam datasets, respectively. Harmformer outperforms previous equivariant transformers and narrows the performance gap between equivariant transformers and convolution-based models.

For completeness, we include the average performance in each benchmark listed in Table \ref{tab:avg-error}. The accuracy on the PCam dataset was slightly unstable, probably due to the characteristics of the dataset. Methods specifically designed for PCam, such as \citep{graham2020}, use extensive augmentation techniques, which were avoided in our case to ensure unbiased results.

\begin{table}[t]
\parbox{.495\linewidth}{
  \caption{Error on mnist-rot-test}
  \label{tab:mnist-rot-test}
  \centering
  \begin{tabular}{llc}
    \toprule
    Model   & Error  & Param. \\
    \midrule
    ResNets-50 \citep{hwang2021} & 57.6\% & \\
    SWN-GCN \citep{hwang2021} & 8.20\% & 2.7M \\
     \ac{hnets} \citep{worrall2017} & 7.11\% & 33k     \\
     H-NeXt \citep{karella2023} & 1.30\%  & 28k \\
     Harmformer & \textbf{0.82\%} & 29.7k\\
    \bottomrule
  \end{tabular}
  }
\parbox{.495\linewidth}{
  \caption{Error on cifar10-rot-test}
  \label{tab:cifar10-rot-test}
  \centering
  \begin{tabular}{llc}
    \toprule
    Model   & Error & Param. \\
    
    \midrule
    ResNets-50 \citep{hwang2021} & 63.90\% & \\
    SWN-GCN \citep{hwang2021} & 49.50\% & 2.7M \\
     H-NeXt\citep{karella2023} & 38.54\%  & 118k \\
     Harmformer & 31.41\% & 118k \\
     Harmformer {\tiny (Large)} & \textbf{29.29\%} & 217k \\
    \bottomrule
  \end{tabular}
  }
\end{table}

\begin{table}[t]
\parbox{.39\linewidth}{
  \caption{Error on rotated MNIST}
  \label{tab:rot-mnist}
  \centering
  \begin{tabular}{llc}
    \toprule
    \ac{sa} model & Error  & Param. \\
    \midrule
    \ac{gsa}\citep{romero2021} & 2.03\%  & 44.7k     \\
    \ac{ge}\citep{xu2023} &  1.99\% & 45k    \\
    Harmformer & \textbf{1.18\%}  & 30k  \\ 
    \midrule
    \ac{cnns} model & & \\
    \midrule
    \ac{gcnn}\citep{cohen2016} & 1.69\% & 73.1k \\
    \bottomrule
  \end{tabular}
  }
\parbox{.20\linewidth}{
  \caption{PCam}
  \label{tab:pcam}
  \centering
  \begin{tabular}{lc}
    \toprule
     Error  & Param. \\
    \midrule
      15.24\%  & 206k     \\
       16.18\% &    \\
      \textbf{12.47\%} & 146k   \\ 
    \midrule
    & \\
    \midrule
     \textbf{10.88\%} & 141k \\
    \bottomrule
  \end{tabular}
}
\parbox{0.40\linewidth}{
  \caption{Average Error}
  \label{tab:avg-error}
  \centering
  \begin{tabular}{ll}
    \toprule
    Dataset & Avg \%$\pm$Std\\
    \midrule
    rotMNIST       & 1.26$\pm$0.055\\
    PCam           & 14.2$\pm0.986$\\
    \midrule
    mnist-rot-test & 0.91$\pm0.142$\\
    cifar-rot-test & 31.9$\pm0.636$\\
    cifar-rot-test {\tiny (Large)} & 29.7$\pm0.203$\\
    \bottomrule
  \end{tabular}
}
\end{table}

\section{Conclusion and Future Work}
The proposed Harmformer is the first transformer model to achieve end-to-end equivariance to continuous rotation and translation in 2D. This was accomplished by designing an equivariant self-attention inspired by harmonic convolution. Along with the novel \ac{sa}, we introduced several layers specifically tailored for equivariance, including linear layers, layer normalization, batch normalization, activations, and residual connections. Our model outperforms previous equivariant transformers, narrowing the performance gap with convolution-based equivariant networks.

We hypothesize that the full potential of transformers may not be realized due to the nature of traditional benchmarks. The 2D equivariant transformers have so far been tested on datasets containing relatively small images that lack global dependencies. Therefore, future research should explore the application of equivariant transformers on larger datasets where, similar to \ac{vit}, they could demonstrate their potential. In addition, the proposed model can be extended to other modalities while maintaining its equivariance properties. For example, the harmonic networks that form the basis of our approach can also be adapted for 3D applications.

\section{Acknowledgments and Disclosure of Funding}
This work was supported by the Czech Science Foundation grant GA24-10069S, by the Ministry of the Interior of the Czech Republic grant VJ02010029 ``AISEE'' and by the grant SVV–2023–260699.

{
\small
\bibliography{neurips_2024}
}

%

\appendix
\include{neurips_appendix}


\end{document}

%% file: neurips_appendix.tex
\newcommand{\rotR}{\mathcal{R}_{\alpha}}
\newcommand{\norm}[1]{\left\lVert#1\right\rVert}
\renewcommand\qedsymbol{$\square$}

\section{Proofs of Harmformer Equivariance}
\label{appendix:equivariance-proofs}
In this section, we systematically formulate the proofs of the \ac{he} property (Definition \ref{def:harmonic-equivariance}) for each layer of the Harmformer. The central concept of the architecture is the handling of three streams corresponding to different rotation orders. By oversimplifying the interactions of rotation orders, two main properties of the harmonic function should be highlighted: 

\begin{enumerate}
    \item Feature maps with the same rotation order can be summed:
\begin{equation*}
    e^{i m \alpha} F_1 + e^{i m \alpha} F_2 = e^{i m \alpha} (F_1 + F_2)
\end{equation*}
    \item Multiplication of feature maps results in the sum of their rotation orders:
\begin{equation*}
 e^{i m_1 \alpha} F_1 \cdot e^{i m_2 \alpha} F_2 = e^{i (m_1 + m_2) \alpha} (F_1 \cdot F_2)  
\end{equation*}
\end{enumerate}
Interactions between these streams occur in layers harmonic convolution or Multi-Head Attention (MSA). Other layers, such as layer normalization, process feature maps of the different rotation order independently. Additionally, some operations, such as batch normalization and activation functions, operate solely on the magnitudes of complex numbers leaving the phase untouched. 
\subsection{Equivariance of Harmonic Convolutions}
\label{appendix:hconv-proofs}
Note that the proof of H-Conv equivariance was originally formulated by \citet{worrall2017}. For the sake of completeness, we have provided a highly simplified version of these proofs. However, we encourage readers to read the more comprehensive work on G-steerable convolution kernels and the theory of steerable equivariant convolution networks in \citep{weiler2023} (Chapters 4-5), which provides a broader perspective and demonstrates the equivalence with \ac{gcnn}.

\begin{lemma}[Rotation of a Harmonic Filter]
\label{lemma:rotation-harmonic-filter}
When the coordinates of a harmonic filter are rotated by an angle $\alpha$, it only changes by a factor $e^{im\alpha}$, where $m$ is the rotation order of the harmonic filter and $\rotR$ is a corresponding 2D rotation matrix.

\begin{proof}
\begin{equation}
\begin{split}
   W_m(\rotR^{-1}\vect{x}) \equiv \tilde{W}_m(r, \theta - \alpha) & = R(r) \cdot e^{-i m (\theta - \alpha)} \\
   & = e^{i m \alpha} \tilde{W}_m(r, \theta) \equiv e^{i m \alpha} W_m(\vect{x}),
\end{split}
\end{equation}
where x is the spatial coordinates. 
\end{proof}
\end{lemma}

Let us denote an input image $I$ that is rotated by the angle $\alpha$ and translated by vector $\vect{t}$ as
\begin{equation}
   \rot{I}_t (\vect{x}) \equiv I(\rotR^{-1} \vect{x} + \vect{t}). 
\end{equation}

\begin{theorem}[Harmonic convolution sums the rotation orders]
When an input image $I$ is rotated by $\alpha$ and translated by $\vect{t}$, the output of a multiple successive harmonic convolution is given by:
\begin{equation}
 \left[ W_{m_1} \circledast W_{m_2} \circledast \cdots \circledast \rot{I}_t \right](\vect{x}) =  e^{i (m_1 + m_2 + \cdots) \alpha}  \left[ W_{m_1} \circledast W_{m_2} \circledast \cdots \circledast I \right](\rotR \vect{x} + \vect{t})
\end{equation}

\begin{proof}
We start with the very first harmonic convolution. 
\begin{align}
    \left[ W_{m_1} \circledast \rot{I}_t \right] (\vect{x}) &= \int_{\mathbb{R}^2} W_{m_1}(\vect{z}) \rot{I}_t (\vect{x} - \vect{z}) d\vect{z} \\
    &= \int_{\mathbb{R}^2} W_{m_1}(\vect{z}) I(\rotR (\vect{x} - \vect{z}) + \vect{t}) d\vect{z} \quad (Eq. 16) \\
    &= \int_{\mathbb{R}^2} W_{m_1}(\rotR^{-1}\vect{z}') I(\rotR \vect{x} - \vect{z}' + \vect{t}) \, d\vect{z}' \quad (\vect{z}' = \rotR \vect{z}) \footnote{Determinant of rotation matrix is 1. } \\
    &= e^{i m_1 \alpha} \int_{\mathbb{R}^2} W_{m_1}(\vect{z}') I((\rotR \vect{x} + \vect{t}) - \vect{z}') d\vect{z}' \quad (\text{Lemma \ref{lemma:rotation-harmonic-filter}}) \\
    &= e^{i m_1 \alpha} \left[ W_{m_1} \circledast I \right] (\rotR \vect{x} + \vect{t})
\end{align} 

Denote the first feature map as $F$, if we roto-translate the input image: 
\begin{equation}
    \rot{F}_t (\vect{x}) \equiv  e^{i m_1 \alpha} F(\rotR \vect{x}  + \vect{t})
\end{equation}

The following harmonic convolution is given by a similar equation.
\begin{align}
    \left[ W_{m_2} \circledast \rot{F}_t \right] (\vect{x}) & = \int_{\mathbb{R}^2} W_{m_2}(\vect{z}) \rot{F}_t(\vect{x} - \vect{z}) d\vect{z} \\
    &=  e^{i m_1 \alpha} \int_{\mathbb{R}^2} W_{m_2}(\vect{z}) F(\rotR \vect{x} - \rotR \vect{z} + \vect{t}) d\vect{z}  \\
    &= e^{i (m_1 + m_2) \alpha} \int_{\mathbb{R}^2} W_{m_2}(\vect{z}') F((\rotR\vect{x}  + \vect{t}) - \vect{z}') d\vect{z}'  \quad (\vect{z}' = \rotR \vect{z}) \\
    &= e^{i (m_1 + m_2) \alpha}  \left[ W_{m_2} \circledast F \right] (\rotR\vect{x}  + \vect{t}) 
\end{align}
Accordingly for all following harmonic convolution layers. 
\end{proof}
\end{theorem}

\subsection{Layers Operating on Magnitudes}
\label{appendix:magnitudes}
This section describes the original \ac{hnets} layers that operate on magnitudes as formulated by \citet{worrall2017}, and introduces our proposed enhancements.

\begin{definition}[$\mathbb{C}$-ReLU]
\label{def:relu}
\begin{equation}
\mathbb{C}\text{-ReLU}_b (X e^{i \theta}) = \text{ReLU}(X + b) e^{i \theta},
\end{equation}
where $X e^{i \theta}$ represents a complex number in exponential form, and $b \in \mathbb{R}$ is a learnable bias parameter of the activation function.
\end{definition}

\begin{definition}[Harmformer $\mathbb{C}$-ReLU]
\begin{equation}
\mathbb{C}\text{-ReLU}_{a, b} (X e^{i \theta}) = \text{ReLU}(a \cdot X + b) e^{i \theta},
\end{equation}
where $X e^{i \theta}$ is a complex number in exponential form, and $a, b \in \mathbb{R}$ are learnable parameters of the activation function.
\end{definition}
Definition \ref{def:relu} uses only the bias parameter $b$. Such an activation function cannot zero out higher values while leaving lower values unaffected. To allow this, our enhanced $\mathbb{C}$-ReLU also incorporates a multiplication by the parameter $a$.
\subsection{Complex Batch Normalization in Harmonic Networks}
\label{sec:complex-bn}

In \ac{hnets}, batch normalization is adapted from its traditional definition. The layer standardizes only the magnitudes of the complex numbers, leaving the phase components unaffected. The $\mathbb{C}$-BN can be formally defined as follows:

\begin{definition}[$\mathbb{C}$-BN]
\begin{equation}
    \mathbb{C}\text{-BN}_{\gamma, \beta}(X e^{i \theta}) = \left(\gamma \left(\frac{X - \mu}{\sqrt{\sigma^2 + \epsilon}}\right) + \beta \right) e^{i \theta} = \text{BN}_{\gamma, \beta}(X) e^{i \theta},
\end{equation}
where $X e^{i \theta}$ represents a complex number in exponential form, and $\gamma, \beta \in \mathbb{R}$ are learnable scaling and shifting parameters, respectively. Here, $\mu$ and $\sigma$ denote the running sample mean and variance, which are estimated during the training phase and fixed during inference.
\end{definition}

However, this formulation can produce negative magnitudes, thus inverting the phase and violating \ac{he}. Therefore in Harmformer we instead use a batch normalization integrated with an activation function, that can be defined as: 
\begin{definition}[Harmformer HBatchNorm + $\mathbb{C}$-ReLU]
\label{appendix:hnormact}
\begin{equation}
 \mathbb{C}\text{-BN-ReLU}_{a,b} (X e^{i \theta})=  \text{ReLU} \left(a \left(\frac{X - \mu}{\sqrt{\sigma^2 + \epsilon}}\right) + b \right) e^{i \theta},
\end{equation}
where $X e^{i \theta}$ represents a complex number in exponential form, and $a, b \in \mathbb{R}$ are learnable scaling and shifting parameters, respectively. Here, $\mu$ and $\sigma$ denote the running sample mean and variance, which are estimated during the training phase and fixed during inference.
\end{definition}
Our formulation uses the ReLU function, which maps $\mathbb{R}$ to $\mathbb{R}^+$, to ensure that changes in magnitudes are always positive. Additionally, by integrating the scaling and shifting parameters \(a\) and \(b\) into batch normalization, the number of learnable parameters is reduced. See section \ref{appendix:norm} for a comparison of different normalization layers. 

\subsection{Discrete Representation}
The normalization layers are the last from the stem stage, as can be seen in Figure \ref{fig:architecture}. For the sake of clarity, we will make the transition to discrete space and focus only on rotation for the following layers, as mentioned in Section \ref{subsec:patches}.
Suppose that the feature maps (stack of patches) $F_m(I) \in \mathbb{C}^{n \times d}$, extracted from the input image $I$, transforms under a rotation of the input as follows
\begin{equation}
F_m(\rot{I}) = e^{i m \alpha} \rot{F_m(I)},
\end{equation}
where $m$ is the rotation order and $\alpha$ is the rotation angle of the image $I$. Here $n$ is the number of patches and $d$ is the dimension of each patch. 

This property implies that $\rot{\cdot}$ is a linear operator, thus for the feature maps the following applies: 
\begin{equation}
    \rot{F_{m_1}(I)} + \rot{F_{m_2}(I)} =  \rot{F_{m_1}(I) + F_{m_2}(I)}
\end{equation}

\begin{equation}
 \label{apeq:linear-mul}
 \rot{F_{m_1}(I)} \cdot \rot{F_{m_2}(I)} =  \rot{(F_{m_1}(I) \cdot F_{m_2}(I)} 
\end{equation}
\subsection{Residual Connection}
\begin{lemma}[\ac{he} of Residual Connections (Lemma \ref{lemma:residual})]
A residual connection between feature maps of the same rotation order, $F_m(I)$ and $F'_m(I)$, preserves \ac{he} property:
\begin{equation}
    F'_m(\rot{I}) + F_m(\rot{I}) = e^{i m \alpha} \rot{(F'_m(I) + F_m(I))}.
\end{equation}
\begin{proof}
    By the properties of harmonic equivariance, we have:
    \begin{align}
        F'_m(\rot{I}) + F_m(\rot{I}) &= (e^{i m \alpha} \rot{F'_m(I)}) + (e^{i m \alpha} \rot{F_m(I)}) \\
        &= e^{i m \alpha} (\rot{F'_m(I)} + \rot{F_m(I)})
    \end{align}
    Since $\rot{\cdot}$ is a linear operator, we can combine the rotated feature maps:
    \begin{align}
        &= e^{i m \alpha} \rot{(F'_m(I) + F_m(I))}
    \end{align}
\end{proof}
\end{lemma}
\subsection{Linear Layers}
\begin{lemma}[\ac{he} of  Linear Layer (Lemma \ref{lemma:linear})]
A linear layer applied to a \ac{he} feature map $F_m(\rot{I}) \in \mathbb{C}^{(hw) \times d_{in}}$ preserves the rotation order $m$. Formally, we have: 
\begin{equation}
    F_m(\rot{I}) W = e^{m i \alpha} \rot{F_m(I) W}, 
\end{equation}
where $W \in \mathbb{C}^{d_{in} \times d_{out}}$ represents a shared weight matrix applied independently over all spatial positions of the input feature map.
\begin{proof}
As the matrix $W$ has a rotation order $m=0$ because it doesn't change under an input rotation. Then the  property comes trivially from Eq~\eqref{apeq:linear-mul}. 
\end{proof}
\end{lemma}
\subsection{Multi-Head Self-Attention}
\begin{lemma}[HE of Layer Norm (Lemma \ref{lemma:layer-norm})]
A feature map $F_m (\rot{I}) \in \mathbb{C}^{(hw)\times d}$ with a rotation order $m$ preserves \ac{he} when normalized by its mean and standard deviation: 
\begin{equation}
   \frac{F_m(\rot{I}) - \mu}{\sigma + \epsilon} = e^{i m \alpha} \frac{\rot{F_m(I)} - \mu}{\sigma + \epsilon},
\end{equation}
where $\mu$, $\sigma$ are the sample means and standard deviations of the original feature maps computed over their spatial dimensions, respectively, and $\epsilon$ is a small constant added for numerical stability.
\begin{proof}
\begin{align}
    \frac{F_m(\rot{I}) - \hat{\mu}}{\hat{\sigma} + \epsilon} &= \frac{F_m(\rot{I}) - \frac{\sum F_m(\rot{I})}{h \cdot w}}{\hat{\sigma} + \epsilon} 
\end{align}
 By the properties of \ac{he}, we express $F_m(\rot{I}) = e^{i m \alpha} F_m(I)$. Sigma of does not change under rotation, because its equal to standard deviation of magnitude in complex numbers. 
\begin{align}
    &=  \frac{e^{i m \alpha}\rot{F_m(I)} - \frac{\sum e^{i m \alpha} \rot{F_m(I)}}{h \cdot w}}{\sigma + \epsilon} \\
    &= e^{i m \alpha} \frac{\rot{F_m(I)} - \frac{\sum \rot{F_m(I)}}{h \cdot w}}{\sigma + \epsilon}
\end{align}
The sum of the feature map is invariant to its rotation.
\begin{align}
    &= e^{i m \alpha} \frac{\rot{F_m(I)} - \mu}{\sigma + \epsilon}
\end{align}
\end{proof}
\end{lemma}
\begin{lemma}[Dot product subtracts rotation orders (Lemma \ref{lemma:dot})]
    Consider two \ac{he} feature maps $Q_{m_1}(\rot{I}) \in \mathbb{C}^{(h w) \times d} $ and $K_{m_2}(\rot{I}) \in \mathbb{C}^{(h w) \times d} $ that represent queries and keys, respectively. The dot product of these feature maps is \ac{he} and has the rotation order $m_1 - m_2$. Formally, we have:
    \begin{equation}
        Q_{m_1}(\rot{I}) \overline{K_{m_2}(\rot{I})^T} = e^{i (m_1 - m_2) \alpha} \rot{Q_{m_1}(I)\overline{K_{m_2}(I)^T}},
    \end{equation}
    where $\overline{K_{m_2}(\rot{I})}^T$ denotes the complex conjugate transpose of $K_{m_2}(\rot{I})$.
    \begin{proof}
        By the properties of harmonic equivariance, we express:
        \begin{align*}
            Q_{m_1}(\rot{I}) &= e^{i m_1 \alpha} \rot{Q_{m_1}(I)}, \\
            K_{m_2}(\rot{I}) &= e^{i m_2 \alpha} \rot{K_{m_2}(I)}.
        \end{align*}
        Taking the complex conjugate transpose of $K_{m_2}(\rot{I})$, we obtain:
        \begin{align*}
            \overline{K_{m_2}(\rot{I})^T} &= \overline{e^{i m_2 \alpha} \rot{K_{m_2}(I)^T}} = e^{-i m_2 \alpha} \rot{\overline{K_{m_2}(I)^T}}.
        \end{align*}
        As is derived from the commutativity of the scalar multiplication with the matrix multiplication.       
        \begin{align*}
            Q_{m_1}(\rot{I}) \overline{K_{m_2}(\rot{I})^T} &= e^{i m_1 \alpha} \rot{Q_{m_1}(I)} e^{-i m_2 \alpha} \rot{\overline{K_{m_2}(I)^T}} \\
            &= e^{i (m_1 - m_2) \alpha} \rot{Q_{m_1}(I)} \rot{\overline{K_{m_2}(I)^T}} \\
            &= e^{i (m_1 - m_2) \alpha} \rot{Q_{m_1}(I) \overline{K_{m_2}(I)^T}}.
        \end{align*}
        This shows that the dot product result is also \ac{he} with a rotation order of $m_1 - m_2$.
    \end{proof}
\end{lemma}

\begin{lemma}[Matrix multiplication sums rotation orders (Lemma \ref{lemma:matmul})]
    Consider a \ac{he} feature map $A_{m_1}(\rot{I}) \in \mathbb{C}^{(hw) \times (hw)}$ representing an attention matrix, and HE feature map $V_{m_2}(\rot{I}) \in \mathbb{C}^{(hw) \times d}$ representing values. The result of their matrix multiplication is HE with a rotation order $m = m_1 + m_2$:
    \begin{equation}
        A_{m_1}(\rot{I}) V_{m_2}(\rot{I}) = e^{i (m_1 + m_2) \alpha} \rot{A_{m_1}(I) V_{m_2}(I)}.
    \end{equation}
    where $\rot{A_{m_1}(I)}$, $\rot{V_{m_2}(I)}$ are feature maps created from unrotated $I$ and rotated afterwards.
    \begin{proof}
        Proof is analogical to Lemma \ref{lemma:dot} By the properties of harmonic equivariance, the HE feature maps $A_{m_1}(\rot{I})$ and $V_{m_2}(\rot{I})$ can be represented as:
        \begin{align*}
            A_{m_1}(\rot{I}) &= e^{i m_1 \alpha} \rot{A_{m_1}(I)}, \\
            V_{m_2}(\rot{I}) &= e^{i m_2 \alpha} \rot{V_{m_2}(I)}.
        \end{align*}
        Multiplying these matrices, we find:
        \begin{align*}
            A_{m_1}(\rot{I}) V_{m_2}(\rot{I}) &= (e^{i m_1 \alpha} \rot{A_{m_1}(I)}) (e^{i m_2 \alpha} \rot{V_{m_2}(I)}) \\
            &= e^{i m_1 \alpha} e^{i m_2 \alpha} \rot{A_{m_1}(I)} \rot{V_{m_2}(I)} \\
            &= e^{i (m_1 + m_2) \alpha} \rot{A_{m_1}(I)} \rot{V_{m_2}(I)}. \\
            &= e^{i (m_1 + m_2) \alpha} \rot{A_{m_1}(I) V_{m_2}(I)}.
        \end{align*}
        This confirms that the product is \ac{he} and preserves the combined rotation order of $m_1 + m_2$.
    \end{proof}
\end{lemma}
\section{Ablation Study and Additional Experiments}
In addition to the experiments in the main text that compare the Harmformer to other methods, we include an ablation study that demonstrates its rotational robustness and explores other architectural choices. We have omitted the PCam benchmark due to computational constraints, as its training time was extremely long. 

\label{appendix:ablation}
\subsection{Ablation: Normalization Layers in Stem Stage (S1)}
\label{appendix:norm}
In Section \ref{appendix:hnormact}, we propose a modification of batch normalization by integrating it with an activation function. Specifically, we first apply batch normalization to the feature magnitudes, followed by a ReLU activation on these normalized values. This approach is more consistent with the original purpose of batch normalization as formulated by \citet{ioffe2015}, which is to standardize the distribution of activations across layers.

To test this novel normalization approach, we replaced our normalization layers in the Harmformer H-Conv block (see Figure \ref{fig:backbone}b) with the original \ac{hnets} batch normalization \citep{worrall2017} followed by a $\mathbb{C}$-ReLU. We also evaluated how our proposed layer normalization used in the encoder block would perform in the H-Conv block.

\begin{figure}[h]
  \centering
  \includegraphics[width=\linewidth]{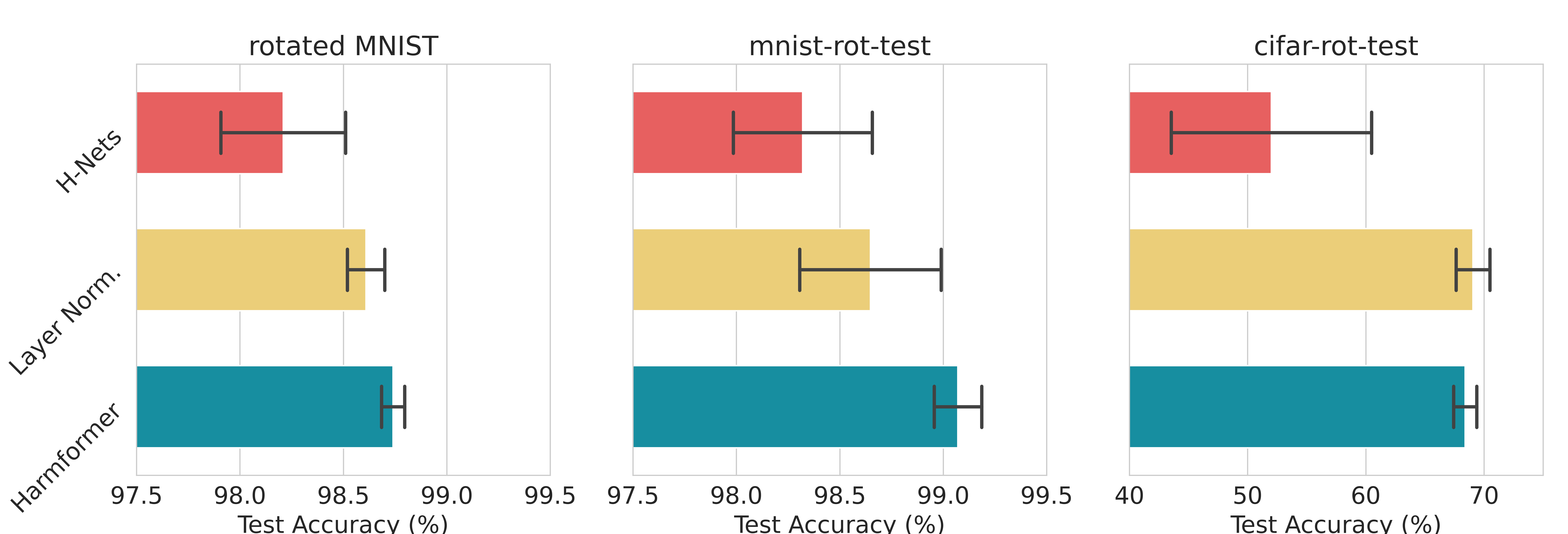}
  \caption{Ablation study on different normalization layers. The rows represent different normalization layers in the H-Conv block. Each plot is aggregated from 5 different runs. The error bars represent the standard deviation.}  
  \label{fig:harmformer-norms}
\end{figure}

The results depicted in Figure \ref{fig:harmformer-norms} show that our proposed normalization (blue bar) outperforms the original \ac{hnets} normalization layer (red bar) across all three benchmarks, significantly reducing variance across different runs. It also exceeds the performance of layer normalization (yellow bar) in the rotated MNIST and mnist-rot-test, although it slightly underperforms in the cifar-rot-test. In addition, the layer normalization was more computationally expensive according to our experiments.

\subsection{Ablation: Mixing Rotation Orders in Self-Attention Mechanism}
\label{appedix:mixing}

\begin{figure}[h]
  \centering
  \includegraphics[width=\linewidth]{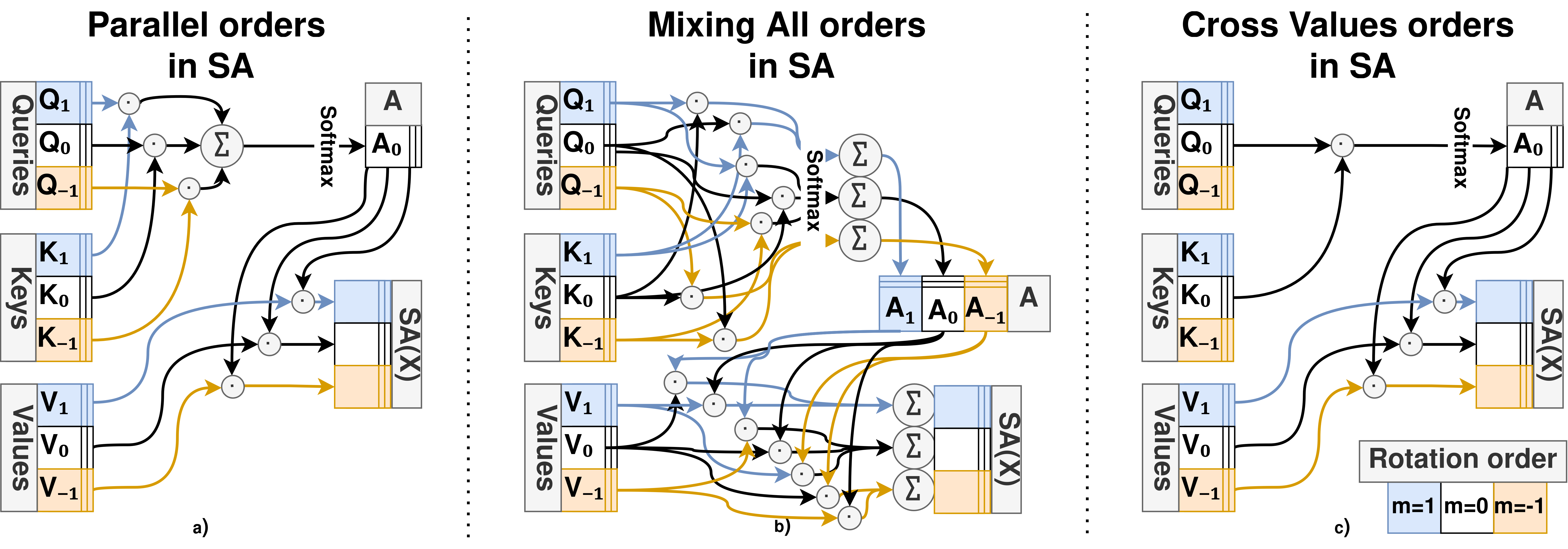}
  \caption{Ablation study on mixing rotation orders within the \ac{sa} mechanism. a) The principle used in Harmformer b) Mixing all possible combinations of queries, keys and values c) Cross Values a method of mixing only values of different rotation orders. }  
  \label{fig:harmformer-mixing}
\end{figure}
As mentioned in the main text (Section \ref{sec:harmformer}), determining how queries, keys, and values should interact based on their rotation order is not intuitive. Therefore, we have extensively tested various configurations and listed the most promising ones in this section. In choosing the final solution, in addition to performance, we focused on the principles that the number of streams should not increase and that the method should not require extensive computation.

In Lemmas~\ref{lemma:dot} and \ref{lemma:matmul}, we demonstrate that the dot product subtracts the rotation orders and matrix multiplication sums the rotation orders. Based on this, we propose several configurations, illustrated in Figure~\ref{fig:harmformer-mixing}. Apart from those mentioned here, we investigated learnable weights for each rotation order combination and different placements of softmax or combinations of these configurations together, but none yielded significant improvements. The final configuration used in Harmformer is shown in Figure~\ref{fig:harmformer-mixing}a. The configuration in Figure~\ref{fig:harmformer-mixing}b allows all possible combinations to produce the three streams (1, 0, -1). The last configuration, Cross Values, illustrated in Figure~\ref{fig:harmformer-mixing}c, uses higher rotation orders only for values. Similarly, we tested Cross Keys and Cross Queries only for keys and queries, respectively.

\begin{figure}[hb]
  \centering
  \includegraphics[width=\linewidth]{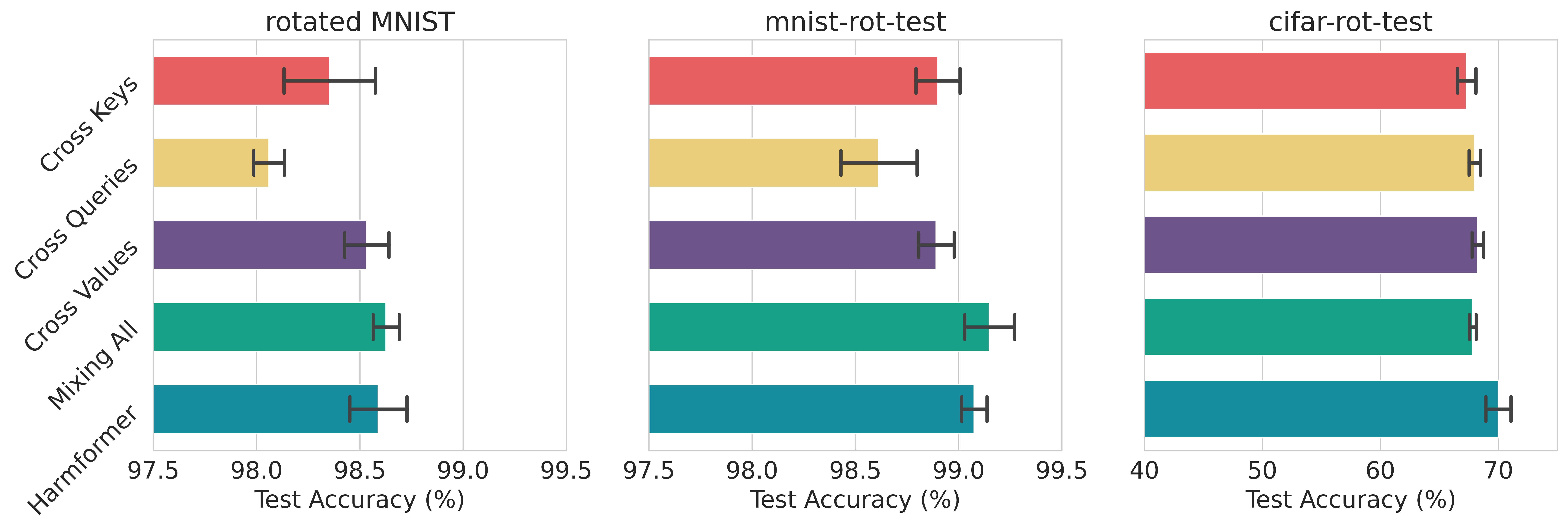}
  \caption{Ablation study on different \ac{sa} mixing configurations. Each plot is aggregated from 5 different runs. The error bars represent the standard deviation.}
  \label{fig:harmformer-mixing-result}
\end{figure}

Figure~\ref{fig:harmformer-mixing-result} presents the performance of these configurations on our benchmarks. The only configuration surpassing Harmformer (Figure~\ref{fig:harmformer-mixing}a) was Mixing All (Figure~\ref{fig:harmformer-mixing}b) in the case of rotated MNIST and mnist-rot-test. Since the performance difference was minor and the computational demands were significantly higher, we did not use Mixing All in our final architecture.

\subsection{Ablation: Relative Positional Encoding (RPE)}
\begin{figure}[h]
  \centering
  \includegraphics[width=\linewidth]{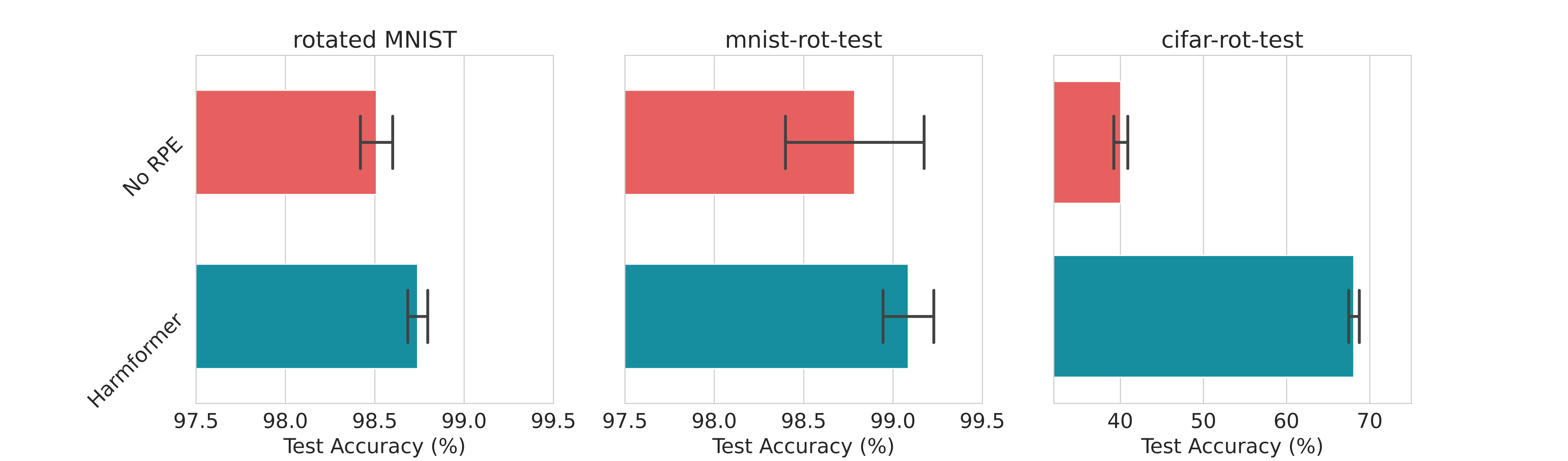}
  \caption{Ablation study on the use of relative circular position encoding (RPE) in the Harmformer. The error bars represent the standard deviation.}  
  \label{fig:harmformer-rpe}
\end{figure}

In Harmformer, we use relative circular encoding similar to those published in iRPE \cite{wu2021}. The encoding is added immediately after calculating the dot product between keys and queries. RPE significantly improves performance, as demonstrated in Figure~\ref{fig:harmformer-rpe}, which shows Harmformer performance with and without RPE.

\subsection{Experiments: Stability of Classification w.r.t. Input Rotation}
\label{appendix:angle-stability}
In these experiments, we investigate the influence of input interpolation errors on performance, following previous invariant models~\citep{khasanova2017, hwang2021, karella2023}. Stability is primarily examined using the invariance benchmarks mnist-rot-test and cifar-rot-test, where the training data consists of non-rotated images, while the test data consists of randomly rotated images. Note that this implies that the training images consist of original sharp images, but the test images contain images with interpolation errors. In contrast, the rotated MNIST dataset contains rotated images in both the training and test sets, resulting in interpolation errors in both sets.

\begin{figure}[hb]
    \centering
    \includegraphics[width=\linewidth]{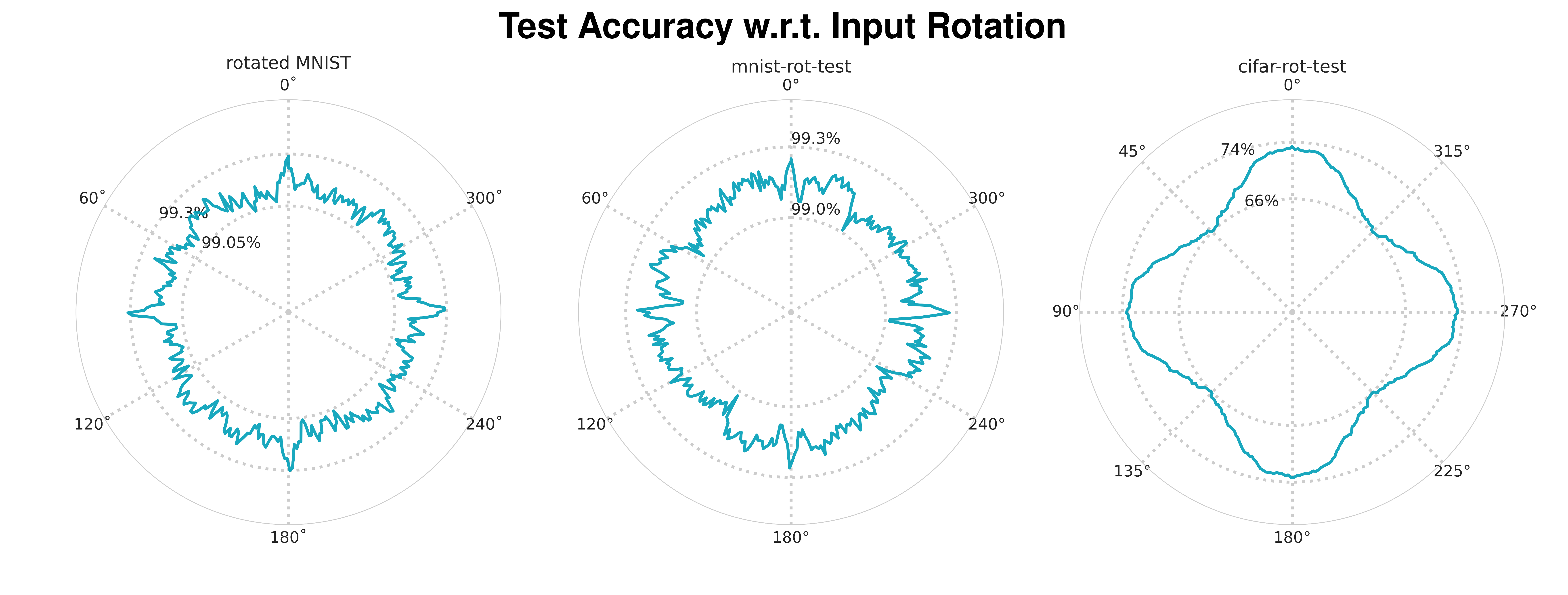}
    \caption{Classification stability with respect to input image rotation. The angle in the circular plots represents the rotation angle of the input image, and the radius represents the test accuracy on specified benchmarks.}
    \label{fig:appendix-stability}
\end{figure}

The test accuracy with respect to the input rotation is shown in Figure~\ref{fig:appendix-stability}.  Since the rotated MNIST dataset does not contain all images rotated by all angles, we use the original MNIST dataset~\citep{lecun1998} for this experiment.  As a result, the test set, and therefore its accuracy, is different from that described in Section \ref{sec:experiments} of the main text.

For the mnist-rot-test, we observe very small oscillations, almost the same as for rotated MNIST. The accuracy reaches maxima at $0^\circ, 90^\circ, 180^\circ,$ and $270^\circ$, where there is no interpolation effect. For the cifar-rot-test, the oscillations are more significant due to the low resolution of the dataset relative to the complexity of the objects, with minima at $45^\circ, 135^\circ, 225^\circ,$ and $315^\circ$, where the interpolation errors are greatest.

For comparison with previous invariant models, we have included Table~\ref{tab:stability-comparison}. For the mnist-rot-test, there is a significant improvement in $\Delta$, which represents the difference in accuracy between the interpolation-free and interpolation-affected images. However, for the cifar-rot-test, the gap between these cases remains almost the same. For the MNIST datasets, the results are almost the same whether training with rotated or unrotated data. This leads to the hypothesis that if the resolution of the dataset matches the complexity of the recognition task, the Harmformer should not suffer from interpolation errors. However, this hypothesis would require further testing on large-scale datasets, which is beyond the scope of this paper.

\begin{table}
  \caption{Test accuracy comparison of the Harmformer and H-NeXt \citep{karella2023} at specific angles of rotation of inputs. }
  \label{tab:stability-comparison}
  \centering
  \begin{tabular}{lcccccc}
    \toprule
    & \multicolumn{3}{c}{mnist-rot-test} & \multicolumn{3}{c}{cifar-rot-test}                 \\
    \cmidrule(r){2-4} \cmidrule(r){5-7}
    Model                       & $0^\circ$     & $45^\circ$ & $\Delta$ & $0^\circ$     & $45^\circ$ & $\Delta$ \\
    \midrule
    H-NeXt~\citep{karella2023}  &  98.9\% &  97.8\%& 1.1\% & 64.5\% & 57.4\% & 7.1\% \\
    Harmformer                  &  99.2\% &  99.1\%& 0.1\% & 73.4\% & 66.1\% & 7.3\% \\
    \bottomrule
  \end{tabular}
\end{table}

\subsection{Experiments: Evaluating the Role of Harmonic Convolutions}
\label{appendix:shallow}
To investigate the importance of the convolutional stem and the encoder, we conducted an experiment using a minimal stem stage with an enlarged attentive field. The purpose of this setup was to ensure that the recognition was not due to convolution alone. This configuration contained only three convolutional layers and a single pooling layer.

The results, as shown in Table \ref{tab:shallow}, indicate that the model performance  remains within the expected error range despite the simplified convolutional stem. This suggested that the encoder plays a crucial role in the final classification. Notably, the inclusion of a single pooling layer significantly enhances the complexity of subsequent attention mechanisms. Due to the increased GPU RAM requirements, this configuration was exclusively tested on the rotated MNIST dataset.
\begin{table}[h]
   \centering
  \caption{Performance comparison of the Harmformer architecture with a shallow stem stage on the rotated MNIST dataset.} 
  \label{tab:shallow}
  \centering
  \begin{tabular}{lccccc}
    \toprule
     Model                 &  SA Input Shape             & Test Error     & Params. \\
    \midrule
     Shallow Stem Stage  &  $32 \times 32 \times  16$  & 1.29\%         & 40k    \\
     Harmformer            &  $16 \times 16 \times  16$  & $1.26\%\pm0.055$ & 30k     \\
    \bottomrule
  \end{tabular}
\end{table}

\section{Experimental Setup}
\label{appendix:experiments}
\subsection{Compute Resource}
Each experiment was run on a single GPU within our shared, small but diverse cluster comprising 17 GPUs. The cluster includes Tesla P100, V100, and A100 models, NVIDIA GeForce RTX 2080 Ti, 3080, 4090, RTX A5000, and a Quadro P5000. Despite its limited size, our setup allowed for flexible and scalable computation using various GPU configurations.
To provide a better overview, Table~\ref{tab:training_time} lists the epoch training time  across each experiment on the NVIDIA GTX 4090.

\begin{table}[h]
\centering
\caption{Training time of one epoch across different benchmarks on the NVIDIA GTX 4090.}
\label{tab:training_time}
\centering
\begin{tabular}{lcccc}
\toprule
GPU Model & mnist-rot-test & cifar-rot-test & rotated MNIST & PCam \\
\midrule
Epoch Training time (mm:ss) & 01:02 & 02:18 & 00:16 & 37:40 \\
Number of Training Samples  &   52k & 42k   &  10k  & 262k  \\
Batch Size                  &   32  &   32  &  32   &  8    \\
\bottomrule
\end{tabular}
\end{table}

\subsection{Computation Complexity w.r.t. Non-Equivariant Convolution and SA Mechanism}
In general, equivariant networks usually due to their properties impose higher computation complexity than their classical counterparts. For example, a single classical convolution has complexity $O(N^2 \cdot n^2)$, where $N \times N$ is the spatial dimension of the output feature map and $n \times n$ is the size of the filter. In contrast, the original G-CNN equivariant to rotation and translation has complexity $O(N^2 \cdot n^2 \cdot |\theta|^2)$, where $\theta$ is the number of elements in the rotation group. Thus, a G-CNN equivariant to 90-degree rotations and translation would have $|\theta| = 4$.

\myparagraph{Harmformer Convolution} In Harmformer stem stage, we use convolution layers similar to \ac{hnets}. This has a complexity of $O(N^2 \cdot n^2 \cdot |o|^2)$, where $|o|$ is the number of rotation orders of the input and output feature maps.

\myparagraph{Harmformer SA mechanism} Classical global \ac{sa} mechanism has a complexity of $O(N^2 \cdot d + N \cdot d^2)$, where $N$ is the number of patches and each patch has dimension $d$. Our \ac{sa} mechanism, as shown in Figure~\ref{fig:encoder}b, adds multiplication by rotation orders $o$ for matrix multiplication and dot product, resulting in a complexity of $O(o \cdot N^2 \cdot d + o \cdot N \cdot d^2)$.

\myparagraph{Additional Computational Considerations} Harmformer operates in the complex domain, where each multiplication requires four times and each addition requires two times more operations than their real counterparts. Additionally, the computational load increases due to upscaling the input and using large convolution kernels, as recommended in H-NeXt \citep{karella2023}. These factors also contribute to the overall complexity of Harmformer.

\subsection{Configurations of Experiments}
\begin{figure}[h]
  \centering
  \includegraphics[width=\linewidth]{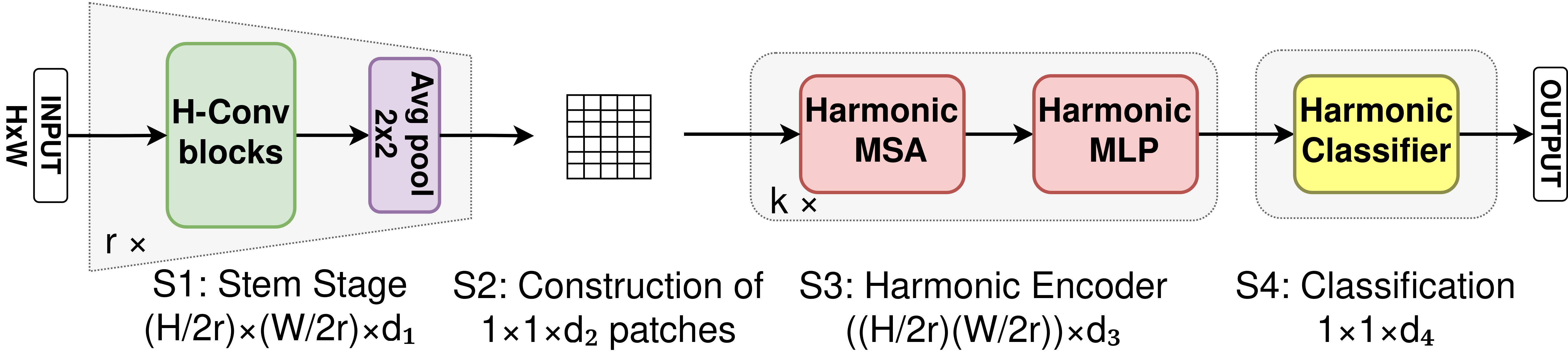}
  \caption{Reproduced from Figure \ref{fig:architecture} to detail the number of parameters for each experiment.} 
  \label{fig:architecture_in_appendix}
\end{figure}

This subsection details the specific configurations of the Harmformer architecture used in the experiments described in Section \ref{sec:experiments}. For convenience, Figure \ref{fig:architecture_in_appendix}, which depicts the complete Harmformer architecture, is included. The parameters for each dataset are enumerated in the following tables: Tables \ref{tab:appendix-mnist-arch} and \ref{tab:appendix-mnist-train} for the MNIST-rot-test and rotated MNIST datasets; Tables \ref{tab:appendix-cifar-arch} and \ref{tab:appendix-cifar-train} for the CIFAR-rot-test dataset; and Tables \ref{tab:appendix-pcam-arch} and \ref{tab:appendix-pcam-train} for the PCam dataset.

\newpage

\begin{table}[h]
\parbox{.50\linewidth}{
  \caption{PCam: Architecture}
  \label{tab:appendix-pcam-arch}
  \centering
  \begin{tabular}{lc}
    \toprule
    Parameter   & Value \\
    \midrule
    Number of S1 Blocks ($r$) &  4 \\
    Convolution per S1 Block  &  2 \\
    Number of S1 Channels  ($d_1$)   & $ \left[4, 8, 16, 32 \right]$ \\
    S1 Channels Dropout       & $\left[ 0, 0.2, 0.3, 0.4  \right]$ \\
    \midrule
    Number of S3 Encoders ($k$)    & 4  \\
    Number of S3 Heads        & 4  \\
    Shape of S3 Patches ($d_3$) & 8 \\ 
    MSA\&MLP Dropout & 0.4 \\
    \bottomrule
  \end{tabular}
  }
\parbox{.50\linewidth}{
  \caption{PCam: Training Settings}
  \label{tab:appendix-pcam-train}
  \centering
  \begin{tabular}{lc}
    \toprule
    Parameter   & Value \\
    \midrule
    Epochs & 100 \\ 
    Batch Size & 8 \\
    Learning Rate & 0.0007 \\ 
    Label Smoothing & 0.1 \\
    Scheduler &  Cosine \\
    Optimizer & AdamW \\
    Weight Decay & 0.01 \\
    Runs & 5 \\
    Input Padding & 0 \\
    \bottomrule
  \end{tabular}
  }
\end{table}

\begin{table}[h]
\parbox{.50\linewidth}{
  \caption{MNIST datasets: Architecture }
  \label{tab:appendix-mnist-arch}
  \centering
  \begin{tabular}{lc}
    \toprule
    Parameter   & Value \\
    \midrule
    Number of S1 Blocks ($r$) &  2 \\
    Convolution per S1 Block  &  2 \\
    Number of S1 Channels  ($d_1$)  & $ \left[8, 16 \right]$ \\
    S1 Channels Dropout       & $\left[ 0, 0  \right]$ \\
    \midrule
    Number of S3 Encoders   ($k$)    & 3  \\
    Number of S3 Heads          & 1  \\
    Shape of S3 Patches ($d_3$) & 16 \\ 
    MSA\&MLP Dropout & 0.1 \\
    \bottomrule
  \end{tabular}
  }
\parbox{.50\linewidth}{
  \caption{MNIST datasets: Training Settings}
  \label{tab:appendix-mnist-train}
  \centering
  \begin{tabular}{lc}
    \toprule
    Parameter   & Value \\
    \midrule
    Epochs & 100 \\ 
    Batch Size & 32 \\
    Learning Rate & 0.007 \\ 
    Label Smoothing & 0.1 \\
    Scheduler & Reduce LR on Plateau \\
    Optimizer & AdamW \\
    Weight Decay & 0.01 \\
    Runs & 5 \\
    Input Padding & 2 \\
    \bottomrule
  \end{tabular}
  }
\end{table}

\begin{table}[h]
\parbox{.50\linewidth}{
  \caption{cifar-rot-test: Architecture}
  \label{tab:appendix-cifar-arch}
  \centering
  \begin{tabular}{lc}
    \toprule
    Parameter   & Value \\
    \midrule
    Number of S1 Blocks ($r$) &  2 \\
    Convolution per S1 Block  &  3 \\
    Number of S1 Channels  ($d_1)$   & $ \left[8, 16 \right]$ \\
    S1 Channels Dropout       & $\left[ 0, 0.1  \right]$ \\
    \midrule
    Number of S3 Encoders ($k$)    & 4  \\
    Number of S3 Heads        & 4  \\
    Shape of S3 Patches ($d_3$) & 8 \\ 
    MSA\&MLP Dropout & 0.2 \\
    \bottomrule
  \end{tabular}
  }
\parbox{.50\linewidth}{
  \caption{cifar-rot-test: Training Settings}
  \label{tab:appendix-cifar-train}
  \centering
  \begin{tabular}{lc}
    \toprule
    Parameter   & Value \\
    \midrule
    Epochs & 200 \\ 
    Batch Size & 32 \\
    Learning Rate & 0.007 \\ 
    Label Smoothing & 0.1 \\
    Scheduler & Cosine \\
    Optimizer & AdamW \\
    Weight Decay & 0.01 \\
    Runs & 5 \\
    Input Padding & 0 \\
    \bottomrule
  \end{tabular}
  }
\end{table}

\newpage
\section{Segmentation experiment: Retina blood vessel segmentation}
\label{sec:segmentation}
To demonstrate the generalizability and scalability of our architecture beyond classification tasks, we introduce Harmformer for retinal blood vessel segmentation using the DRIVE dataset \cite{staal2004}. The DRIVE is binary segmentation task, where the goal is to extract retinal blood vessels from an RGB image.

The dataset contains 262,080 samples for training and 65,520 samples for validation, similar to the settings of \citep{bekkers2018}.  Each sample consists of an input image of size $3 \times 64 \times 64$ and a target segmentation mask of size $64 \times 64$. These samples were generated from 17 training images and 3 validation images, each of which is $768 \times 584$ pixels and represents a different patient.

\begin{figure}[h]
    \centering
    \includegraphics[width=\linewidth]{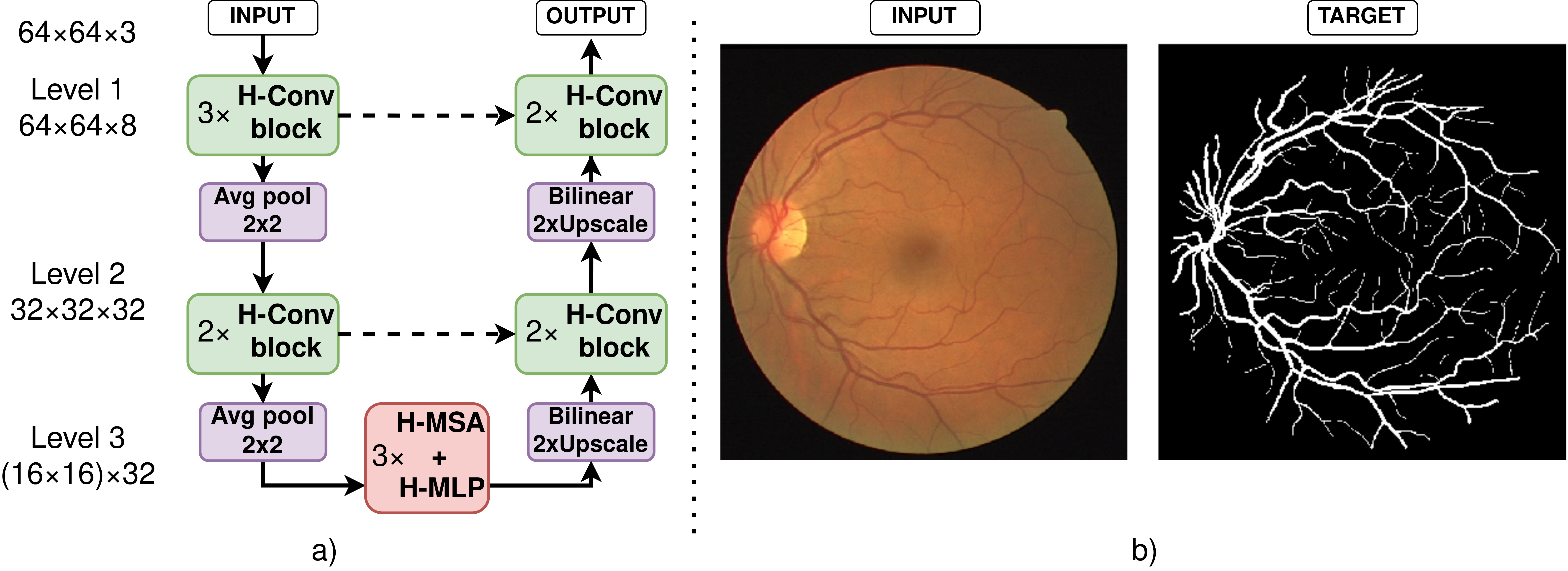}
    \caption{(a) Diagram of the Harmformer architecture for image segmentation. (b) Example of an image from the DRIVE dataset: the RGB image and the target segmentation mask.}
    \label{fig:harmunet}
\end{figure}

To use Harmformer as an image-to-image model, we adopt a U-Net~\cite{ronneberger2015} architecture in Fig~\ref{fig:harmunet}a. Unlike our classification models (Section~\ref{sec:experiments}), this model processes the images at their original resolution, without any upscaling before they enter the network. For the output, we use only the magnitude of the final feature maps. To merge the hidden features (channels) into a single output layer, we apply a standard 2D convolution layer at the end.

We trained the U-Net Harmformer for 20 epochs with the AdamW optimizer, a learning rate of 0.001 and 64 batch size. For augmentation, we used horizontal and vertical flipping, color jitter, and auto-contrast. We ran 4 different experiments with different seeds.

The results are shown in Table~\ref{tab:auc}, using the area under the receiver operating characteristic curve (AUC) as the evaluation metric. For completeness, we have also included the performance of \ac{gcnn} and the current state-of-the-art model FR-UNet~\cite{liu2022}. As expected, these results are consistent with the findings in the paper, with Harmformer slightly underperforming compared to equivariant convolution architectures. Nevertheless, we show that our architecture is versatile and can also be applied to non-classification tasks. 

\begin{table}[h]
   \centering
  \caption{AUC for DRIVE segmentation \cite{staal2004}.} 
  \label{tab:auc}
  \centering
  \begin{tabular}{lcc}
    \toprule
     Model                      &  AUC               & Equivariant model\\
    \midrule
     Harmformer                 &  $0.9746\pm0.0002$  &   \cmark   \\
     G-CNNs \cite{bekkers2018}  &  $0.9784\pm0.0001$  &   \cmark   \\ 
     FR-UNet \cite{liu2022}     &  $0.9889$           &   \xmark   \\ 
    \bottomrule
  \end{tabular}
\end{table}

\newpage

\section{Differences Between 2D and 3D Equivariant Transformers}
While 2D equivariant transformers \citep{romero2021, xu2023} have been relatively understudied, 3D equivariant transformers \citep{assaad2023, liao2024, liao2023, he2021, hutchinson2021, fuchs2020} have received more attention. In this section, we aim to highlight the key differences that make the 2D case unique, and compare Harmformer with the most closely related $SE(3)$-Transformer, which operates in 3D but also uses steerable basis representations.

An important distinction lies in the nature of the input data, which directly influences the transformer architecture. While 2D datasets typically consist of dense pixels with highly correlated neighborhoods, 3D equivariant datasets, often represented as graphs or point clouds, tend to be sparse. In 3D, neighboring elements can vary significantly; for example, in molecular graphs \citep{liao2024,liao2023}, atoms can fulfill entirely different roles within the structure.

\myparagraph{Patches} The properties of the input data determine how to prepare patches in an equivariant manner. In the case of the $SE(3)$-Transformer, each node of the graph can be directly treated as a patch, eliminating the need for a stem stage.  For Harmformer, on the other hand, it is necessary to aggregate low-level correlated data into a higher-level representation. Additionally, the classical ($16 \times 16$) ViT~\citep{vaswani2017} grid  cannot be used, as discussed in Section~\ref{sec:equivariance}. Therefore, we employ a convolutional stem stage, where the convolution kernels are expressed using circular harmonics to maintain equivariance.

This reliance on harmonic representations is a common feature between Harmformer and the $SE(3)$-Transformer. While Harmformer uses circular harmonics, the $SE(3)$-Transformer uses spherical harmonics. Both approaches leverage steerable basis functions~\citep{freeman1991}, which are widely used in equivariant networks~\citep{weiler2018a, weiler2018b, kondor2018}. These steerable bases change predictably under rotation, allowing the effects of rotation to be effectively neutralized—via phase shifts in circular harmonics and via the Wigner-D matrix in spherical harmonics. It is important to note that the use of steerable bases predates both transformers, as shown in~\citep{freeman1991, weiler2018a, kondor2018}.

\myparagraph{Queries, Keys, and Values} In Harmformer, queries (Q), keys (K), and values (V) are generated independently from individual patches through a linear layer, we proposed in Section~\ref{subsec:patches}. In contrast, the $SE(3)$-Transformer creates them by applying convolutions across points (i.e., patches), using steerable spheres that aggregate information from the local neighborhood. 

\myparagraph{Attention} The $SE(3)$-Transformer focuses exclusively on invariant attention (type-0) and applies only local attention. In contrast, Harmformer explores multiple strategies for mixing attention and values of various orders (types), while performing global attention across the entire image. Additionally, Harmformer introduces an equivariant layer normalization at the beginning of the attention layer, while the $SE(3)$-Transformer does not use any layer normalization. Other minor distinctions include Harmformer's use of an improved activation function and relative embeddings.